%% file: main.tex
\documentclass[10pt,twocolumn,letterpaper]{article}

\usepackage[pagenumbers]{cvpr} 

\input{preamble}
\definecolor{cvprblue}{rgb}{0.21,0.49,0.74}
\usepackage[pagebackref,breaklinks,colorlinks,allcolors=cvprblue]{hyperref}
\usepackage{graphicx}
\usepackage{wrapfig}
\usepackage{multirow}
\usepackage{booktabs}
\usepackage[table]{xcolor}
\usepackage{relsize}
\usepackage[most]{tcolorbox}
\def\paperID{****} 
\def\confName{CVPR}
\def\confYear{2026}

\definecolor{mygreen}{RGB}{88, 142, 50}
\definecolor{mygray}{rgb}{0.7,0.7,0.7}
\definecolor{mygray2}{rgb}{0.5,0.5,0.5}

\title{VisMem: Latent Vision Memory Unlocks Potential of Vision-Language Models}
\author{
   Xinlei Yu$^{1}$ $\quad{}$ Chengming Xu$^{2}$ $\quad{}$ Guibin Zhang$^{1}$ $\quad{}$ Zhangquan Chen$^{3}$ $\quad{}$ Yudong Zhang$^{5}$ \\
   Yongbo He$^{4}$ $\quad{}$ Peng-Tao Jiang$^{6}$ $\quad{}$ Jiangning Zhang$^{4}$ $\quad{}$ Xiaobin Hu$^{1}$\thanks{Corresponding authors.} $\quad{}$ Shuicheng Yan$^{1}$\\
\small$^1$National University of Singapore \quad 
  $^2$Fudan University\quad 
  $^3$Tsinghua University \quad
  $^4$Zhejiang University  \\ 
\small$^5$University of Science and Technology of China \quad 
  $^6$vivo \\
}

\begin{document}
\maketitle

\input{sec/0_abstract}

\input{sec/1_intro}
\input{sec/2_related}

\input{sec/3_methodology}
\input{sec/4_experiment}

\input{sec/5_conclusion}
\nocite{gu2025thinkmorph,sun2025latent,li2025finecir,yu2025visual2,he2024multi,chen2025sifthinker,bi2025llava,wang2025ascd,shen2025align,chen2025reasoning,zhao2025cot,dong2025insight}
{
    \small
    \bibliographystyle{ieeenat_fullname}
    \bibliography{main}
}

\input{sec/X_suppl}

\end{document}

%% file: sec/0_abstract.tex
\begin{abstract}
Despite the remarkable success of Vision-Language Models (VLMs), their performance on a range of complex visual tasks is often hindered by a ``visual processing bottleneck": a propensity to lose grounding in visual evidence and exhibit a deficit in contextualized visual experience during prolonged generation. 
Drawing inspiration from human cognitive memory theory, which distinguishes short-term visually-dominant memory and long-term semantically-dominant memory, we propose \textbf{VisMem}, a cognitively-aligned framework that equips VLMs with dynamic latent vision memories, a short-term module for fine-grained perceptual retention and a long-term module for abstract semantic consolidation.
These memories are seamlessly invoked during inference, allowing VLMs to maintain both perceptual fidelity and semantic consistency across thinking and generation. 
Extensive experiments across diverse visual benchmarks for understanding, reasoning, and generation reveal that VisMem delivers a significant average performance boost of 11.0\% relative to the vanilla model and outperforms all counterparts, establishing a new paradigm for latent-space memory enhancement. The code will be available: \href{https://github.com/YU-deep/VisMem.git}{https://github.com/YU-deep/VisMem.git}.
\end{abstract}

%% file: sec/1_intro.tex
\vspace{-16pt}
\section{Introduction}
\label{sec:intro}
Visual-Language Models (VLMs) have demonstrated impressive capabilities in visual understanding, reasoning and generation~\cite{li2024survey,su2025thinking}. Latest flagship models, both closed-sourced~\cite{gpt5,gemini,anthropic2025claude} and open-sourced~\cite {bai2025qwen2,team2025kimi,an2025llava,wang2025internvl3,vteam2025glm45v}, represent a significant leap towards a general-purpose intelligent model that can both perceive and think about the visual world. Despite their success, VLMs still face significant inherent challenges when tackling complicated tasks that require advanced visual abilities, such as fine-grained perception, multi-step reasoning, or maintaining fidelity over long generative sequences~\cite{ghosh2024exploring,huang2025visfactor}. 
A fundamental limitation stems from the pervasive propensity, exhibited during deep autoregressive decoding, toward a deficit in visual memory, which prioritizes accumulated textual context over the initial visual evidence and lacks visual semantic knowledge~\cite{zhou2024rethinking,sun2025mitigating}. It manifests as a ``visual processing bottleneck" that impairs performance in fine-grained visual understanding, efficient reasoning, and robust generation.

\begin{figure*}[t]
  \centering
    \includegraphics[width=0.9\linewidth]{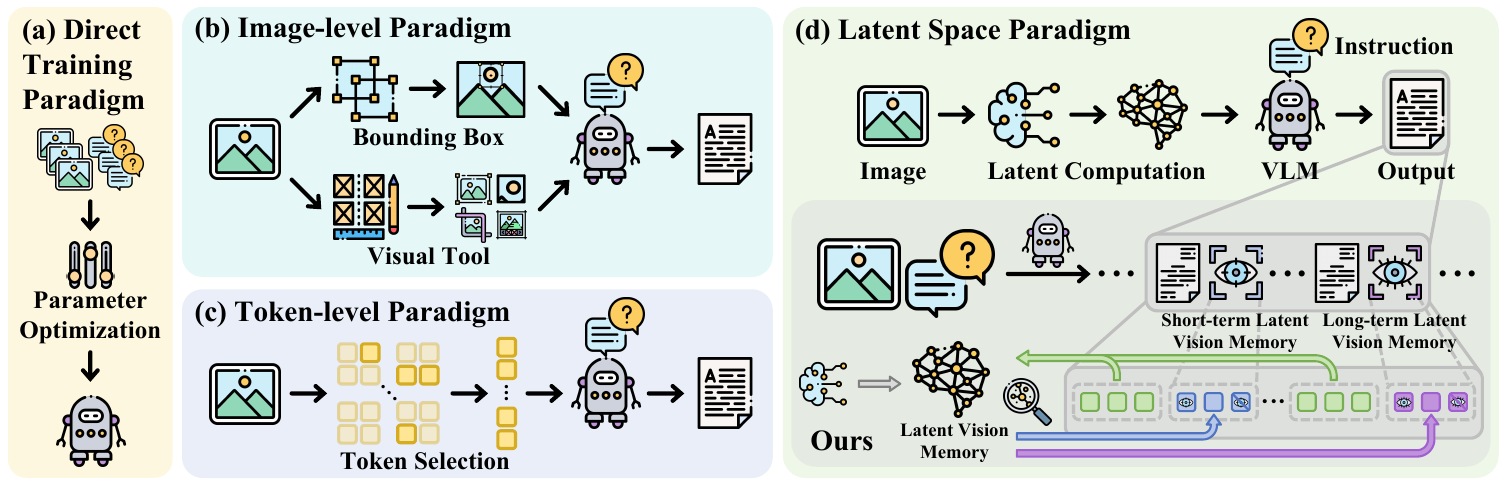}
    \vspace{-3mm}
    \caption{Four primary paradigms for enhancing visual capabilities: (a) the direct training paradigm, (b) the image-level paradigm, (c) the token-level paradigm, and (d) the latent space paradigm. Our VisMem belongs to the last one, featuring latent vision memory.}
    \label{fig:comparison}
    \vspace{-10pt}
\end{figure*}

Prior efforts to overcome this limitation have explored several distinct strategic axes, which can be primarily categorized into four paradigms, as illustrated in \cref{fig:comparison}. One intuitive paradigm is the \textbf{(a) direct training paradigm}, which optimizes model parameters via fine-tuning or reinforcement learning~\cite{liu2025visual,shen2025vlm,huang2025vision,wang2025perception}. This relatively brute-force approach often sacrifices generalization for task-specific performance, leading to catastrophic forgetting. Another axis concerns the representation space of the intervention, \textbf{(b) image-level paradigm}, operating in the pixel space by explicitly synthesizing new visual inputs, which offers image-level thinking but at a prohibitive computational cost~\cite{zheng2025deepeyes,fan2025grit,hu2024visual,su2025pixel,li2025imagine,su2025openthinkimg}. Conversely, \textbf{(c) token-level paradigm} constrains operations to visual tokens, which is more efficient but fundamentally non-generative, limiting the model to merely re-surfacing what it has already encoded~\cite{lei2024scaffolding,gao2025interleaved,chen2025mint,yu2025introducing}. Recently, a promising direction lies in the \textbf{(d) latent space paradigm}, which introduces continuous latent contexts in the sequential inference process. Unfortunately, existing latent space methods either rely solely on the language space~\cite{hao2024training,xu2025softcot,shen2025codi,li2025seek,zhang2025memgen} or require auxiliary visual data~\cite{yang2025machine}, limiting their application in VLMs.

To overcome this problem, we resort to cognitive psychology, specifically the \textit{Dennis Norris Theory}~\cite{norris2017short}: 
\begin{tcolorbox}[colframe=black!50, colback=cvprblue!8, boxrule=1.5pt, arc=2mm, top=4pt, bottom=4pt, left=4pt, right=4pt,  boxsep=1pt]
\raisebox{-0.2\baselineskip}{\includegraphics[height=1\baselineskip]{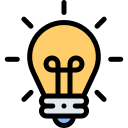}} \textit{ Short-term memory and long-term memory are two distinct storage systems that can be modeled on their neural underpinnings, the former  is governed by vision, while the latter holds sway over abstract semantics.}
\end{tcolorbox}
\noindent While this cognitive theory reveals the essence of human cognition, it can be smoothly translated into an architectural principle of VLMs: short-term memory is visually-dominant, enhancing perception of the current visual scenes, while long-term memory is semantically-dominant, providing generalized knowledge and contextualized semantic, completing the full cognitive chain.

Based on such inspiration, we propose VisMem, a novel and cognitively-aligned framework that systematically incorporates short- and long-term latent vision memory into VLMs. VisMem functions by non-intrusively extending the vocabulary of VLMs with special tokens that trigger on-demand latent vision memory invocation during autoregressive generation. Upon generating an invocation token, a lightweight query builder assesses the hidden states, which contains the current multi-modal cognition, to formulate a contextual-aware query which is then dispatched to one of two specialized, lightweight memory formers: short-term memory former that generates latent tokens encoding fine-grained, perceptual evidences of current visual inputs; long-term memory former that synthesizes tokens representing abstract, high-level semantic knowledge. These generated latent memory tokens are seamlessly inserted into the generation stream, enriching the contexts and enabling it to output with a seamless integration of detailed visual information and generalized semantic knowledge.

With a two-stage training paradigm based on reinforcement learning tailored for our proposed framework, the model learns to first generate effective memory contents, based on which the optimal patterns for invoking the memory is then learned. Our extensive experiments across a wide range of benchmarks spanning visual understanding, reasoning, and generation demonstrate that our approach can substantially enhance the comprehensive visual capabilities on various base models, while also improving cross-domain generalization and mitigating the problem of catastrophic forgetting. Our contributions are listed as follows:

\begin{itemize}
    \item We propose a novel paradigm to proactively harness vision memory, alleviating the ``visual processing bottleneck" and augmenting advanced visual capabilities.
    \item We propose a short- and long-term latent vision memory system with distinct purposes and mechanisms, which is analogous to the cognitive psychology.
    \item We propose a dynamic memory invocation mechanism for seamlessly invoking and inserting latent memory tokens into the autoregressive inference process. 
    \item We evaluate the framework on extensive benchmarks, showcasing significant improvements in advanced visual capacities, cross-domain generalization, catastrophic forgetting mitigation, and compatibility across base models. 
\end{itemize}

%% file: sec/2_related.tex
\section{Related Work}
\subsection{Visual Capacities Enhancement}
As demonstrated in \cref{fig:comparison}, existing methods to alleviate ``visual processing bottleneck" of VLMs broadly fall into four main categories: 
\textbf{(a) direct training paradigm}, which directly optimizes model parameters for target visual tasks, as in SFT, Visual-RFT~\cite{liu2025visual}, VLM-R1~\cite{shen2025vlm}, Vision-R1~\cite{huang2025vision}, and PAPO~\cite{wang2025perception}. Nonetheless, these methods suffer from catastrophic forgetting, specifically manifested as the degradation of general capabilities and over-specialization in specific visual cognition tasks~\cite{zhou2023learning,yu2024boosting};
\textbf{(b) image-level paradigm}, which either leverages bounding boxes to denote visual evidence, represented by methods as Visual CoT~\cite{shao2024visual}, DeepEyes~\cite{zheng2025deepeyes}, SpatialVTS~\cite{liang2025enhancing}, VGR~\cite{wang2025vgr}, and GRIT~\cite{fan2025grit}, or externally generate the iterative visual inputs via predefined tools, as seen in Sketchpad~\cite{hu2024visual}, VPRL~\cite{xu2025visual}, PyVision~\cite{zhao2025pyvision}, OpenAI o3~\cite{openai2025o3}, PixelReasoner~\cite{su2025pixel}, MVoT~\cite{li2025imagine}, and OpenThinkImg~\cite{su2025openthinkimg}. Nevertheless, modifying visual inputs incurs extremely high computational costs, accompanied by high latency and reliance on external tools and concretized images;
\textbf{(c) token-level paradigm}, which select original representations and cannot modify visual evidences, thus restricted by insufficiently refined information and suboptimal selection strategies, as in ICoT~\cite{gao2025interleaved}, MINT-CoT~\cite{chen2025mint}, SCAFFOLD~\cite{lei2024scaffolding}, LLaVA-AURORA~\cite{bigverdi2025perception}, VPT~\cite{yu2025introducing}, Chameleon~\cite{team2024chameleon}, 
\textbf{(d) latent space paradigm}, which employs latent states to optimize autoregressive generation, but its focus remains on pure language models, \textit{e.g.}, Coconut~\cite{hao2024training}, MemGen~\cite{zhang2025memgen}, LatentSeek~\cite{li2025seek}, SoftCoT~\cite{xu2025softcot}, CODI~\cite{shen2025codi}. Although Mirage~\cite{yang2025machine} attempts to construct a latent vision space, requiring substantial manually labeled images. Our VisMem also belongs to this paradigm, but differs from existing methods by integrating latent vision memory within generation processes, characterized by a short and long memory system.

\subsection{Memory Empowerment}
Another mechanism closely tied to our approach involves endowing models with memory functionality. One intuitive strategy entails directly optimize models on prior trajectories, exemplified by~\cite{shen2024memory,zhang2025g,fu2025agentrefine}, or to store them into the external memory repositories~\cite{tack2024online,wang2024wise}. Besides, some models inject persistently stored, retrieval-augmented knowledge from external environments, such as Expel~\cite{zhao2024expel} and MemoryBank~\cite{zhong2024memorybank}, others, such as SkillWeaver~\cite{zheng2025skillweaver} and Alita~\cite{qiu2025alita}, distill prior knowledge as reusable tools. Currently, latent memory, as an implicit memory representation with better cross-domain generalization, efficiently encodes deep semantic associations, including M+~\cite{wang2025m} and MemGen~\cite{zhang2025memgen}. 
Nevertheless, these memory paradigms fail to ideally accommodate visual information, which manifests as a continuous, high-dimensional perceptual input. Consequently, the exploration of efficient visual memory mechanisms remains a largely uncharted territory. Thus, we propose a more human-aligned latent vision memory paradigm.


%% file: sec/3_methodology.tex
\section{Methodology}

\subsection{Preliminary}
\noindent \textbf{Problem Formulation.} Based on the interaction process of VLMs, we formulate the problem and introduce the notations used. We first define a policy model $\mathcal{P}$, which is powered by a base VLM. Given a visual task to be solved, feeding a instruction-vision pair $\left(I,V\right)$ sampled from a task distribution $\mathcal{D}$, the policy model unfolds a corresponding trajectory $\tau$ at a timestep $t$, including pairs of current state $s_t$ of the environment and the action $a_t$ performed by the model. Here, the state of the environment includes textual contexts and visual observations. Internally, the action is generated sequentially by the token-by-token autoregressive decoding of the model, yielding the output token sequence $\{x_{t,1},x_{t,2},\dots,x_{t,l}\}$. The generation of \textit{i-th} output token $x_{t,i}$ could be presented as:
\vspace{-3pt}
\begin{equation}
    x_{t,i} \sim \mathcal{P}\left(\cdot \mid s_t, x_{<i}\right),
\end{equation}
where the prediction is conditioned on the current environment state and previously generated tokens. To endow the model with vision memory, a vision memory system $\mathcal{M}$ is adhered to the policy model, thus, the objective is to optimize the memory-enhanced model jointly and to maximize its expected performance:
\vspace{-3pt}
\begin{equation}
    \max _{\mathcal{P}, \mathcal{M}} \mathbb{E}_{\left(I,V\right)\sim \mathcal{D},\tau \sim \left (\mathcal{P,M}\right)}[S(\tau)],
\end{equation}
where $S\left(\cdot\right)$ denotes the quantifiable performance results, \textit{e.g.}, accuracy or signal from a reward model. 

\begin{figure*}[t]
  \centering
    \includegraphics[width=0.885\linewidth]{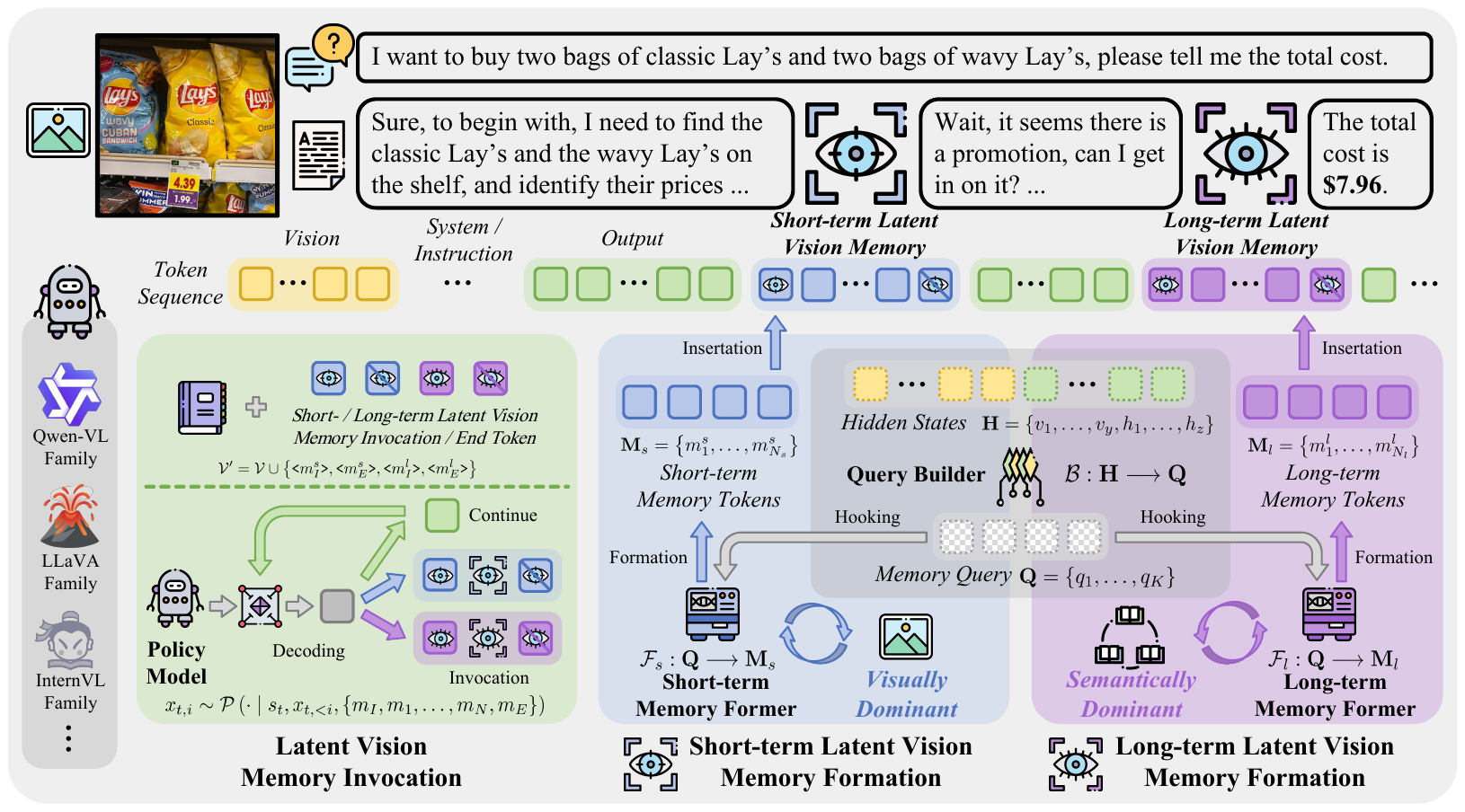}
    \vspace{-3mm}
    \caption{The overview of our proposed VisMem.}
    \label{fig:overview}
    \vspace{-10pt}
\end{figure*}

\vspace{2pt}
\noindent \textbf{Motivation.} Building on the \textit{Dennis Norris Theory}~\cite{norris2017short}, which aligns with contemporary models of human memory, the coordinated operation of short- and long-term visual memories surmounts the ``visual processing bottleneck". Short-term latent visual memory maintains fine-grained detail for immediate use and is thus visually dominant; by contrast, long-term latent visual memory abstracts across experiences to enable flexible reuse and is therefore semantically dominant. Taking the task illustrated in \cref{fig:overview} as a case in point, ``find the classic Lay's on the shelf" entails the deployment of short-term vision memory, retaining visual details for immediate perceptual demands, while ``get in the promotion" triggers generalized semantic knowledge about the ``promotion label" acquired from historical scenarios, which is grounded in long-term latent memory, to facilitate the comprehension of the task-based sight. Existing paradigms for enhancing visual capabilities fail to adequately consider vision memory, thus, our VisMem proposes a latent memory method to bridge this gap. More theoretical foundations are in Appendix~{\color{cvprblue}{6}}.

\vspace{2pt}
\noindent \textbf{Memory System.} Based on previous contents, the task could be further disassembles into two main interactive parts: 
\textit{\textbf{memory invocation}} (\cref{sec:memory_invocation}): related to ``where and how to invoke the short- or long-term vision memory"; \textit{\textbf{memory formation}} (\cref{sec:memory_formation}): related to ``what content should the short- or long-term vision memory convey". Additionally, these two decomposed processes interact closely with each other, with distinct priorities and objectives, requiring a meticulously designed \textit{\textbf{training recipe}} (\cref{sec:training}).

\subsection{Memory Invocation}
\label{sec:memory_invocation}

As illustrated in \cref{fig:overview}, our latent vision memory invocation strategy largely aligns with the standard generation pipeline of VLMs, thereby preserving their robust fundamental visual capabilities. Typically, VLMs generate rationales and answers; however, such pure text sequences lack the granularity to capture fine-grained visual perceptions and semantics, which poses challenges to accurate visual understanding, reasoning, and generation. This limitation arises because during inference, VLMs tend to prioritize accumulated textual context over visual evidence, a phenomenon particularly pronounced in long sequences \cite{ghosh2024exploring,yu2025visual,yin2025clearsight,huang2025visfactor}. 
To address this, we extend the vocabulary $\mathcal{V}$ of VLMs by incorporating four additional memory-operation tokens, resulting in $\mathcal{V}^\prime = \mathcal{V} \cup \left\{ \texttt{<}m^s_I\texttt{>}, \texttt{<}m^s_E\texttt{>}, \texttt{<}m^l_I\texttt{>}, \texttt{<}m^l_E\texttt{>} \right\}$. Here, $\texttt{<}m_I\texttt{>}$ and $\texttt{<}m_E\texttt{>}$ form paired invocation and end tokens, where the superscripts $s$ and $l$ denote short- or long-term memory, respectively.
Specifically, we register these as indivisible special tokens in the tokenizer and enlarge the embedding matrix from $\mathbb{R}^{\mid\mathcal{V}\mid\times d}$ to $\mathbb{R}^{\left(\mid\mathcal{V}\mid+4\right)\times d}$, where $d$ is the dimension of the model. Furthermore, we initialize the embeddings of the invocation tokens ($\texttt{<}m^s_I\texttt{>}$ and $\texttt{<}m^l_I\texttt{>}$) using the embedding vector of a delimiter token with small perturbations, and update these embeddings during training to facilitate faster convergence. The two end tokens ($\texttt{<}m^s_E\texttt{>}$ and $\texttt{<}m^l_E\texttt{>}$) are treated as structural markers; they are initialized analogously with a lower learning rate. In practice, we also employ constrained decoding to encourage well-formed invocation-end pairs.

Specifically, the latent vision memory invocation tokens function as triggers for initiating memory insertion, based on the continuous internal cognitive states. During autoregressive generation (see \cref{eq:autograssive}), upon the output of an invocation token, the memory former immediately initiates the latent vision memory formation procedure:
\vspace{-3pt}
\begin{equation} 
x_{t,i}  \rightarrow
\begin{cases}
\text{invocation}, &x_{t,i}\in \left\{\texttt{<}m^s_I\texttt{>},\texttt{<}m^l_I\texttt{>}\right\}\\
\text{continue}, &otherwise
\end{cases}.
\end{equation}
The resulting latent vision memory, whether short- or long-term as dictated by the specific token type, is subsequently inserted right after the already output invocation token. Following this insertion, the corresponding end token for short ($\texttt{<}m^s_E\texttt{>}$) or long memory ($\texttt{<}m^l_E\texttt{>}$) is automatically appended to resume token-by-token decoding:
\vspace{-3pt}
\begin{equation}
    x_{t,i} \sim \mathcal{P}\left(\cdot \mid s_t, x_{t,<i},\left\{m_I,m_1,\dots,m_N,m_E\right\}\right).
    \label{eq:autograssive}
\end{equation}

\subsection{Memory Formation}
\label{sec:memory_formation}
To activate the vision memory capability of VLMs, we integrate two memory components: short-term vision memory, which encodes rich visual evidence, and long-term vision memory, which primarily encodes high-level, knowledge-based visual pertinent semantics, without modifying the core VLM and damaging general abilities. This integration leverages short-term memory to enhance advanced visual perception and comprehension, while long-term memory enables the generalization of semantic experiences during reasoning, thus comprehensively enhancing the overall visual performance. As illustrated in \cref{fig:overview}, the memory formation process hinges on two core components: a query builder $\mathcal{B}$, which is responsible for generating queries to hook memory; and memory formers $\mathcal{F}_s$ and $\mathcal{F}_l$, which are dedicated to constructing latent visual memories.

\vspace{2pt}
\noindent \textbf{Query Builder.} Through this process, we transform hidden states incorporating current cognition into a more efficient and accurate memory query. Initially, we instantiate a lightweight transformer encoder denoted as $\mathcal{B}$ and a learnable memory query $\mathbf{Q}_{init} = \{q_1, \dots, q_K\}$, where $K$ represents the length of the query sequence and each $q\in \mathbb{R}^{d}$. Given the state at a particular time, $\mathcal{B}$ encodes the query sequence based on internal visual and contextual hidden states to retrieve the corresponding latent memory contents. 
During each invocation, as the policy model generates the current output token sequence, \textit{i.e.}, the token sequence starting from the initial position or from the end of the previous invocation, it accordingly produces a sequence of hidden state vectors $\left\{h_1, \dots, h_z\right\}$. Similarly, visual encoder produces visual hidden state vectors $\left\{v_1, \dots, v_y\right\}$. Thus, the combination of them $\mathbf{H} = \{v_1, \dots, v_y,h_1, \dots, h_z\}\in\mathbb{R}^{\left(y+z\right)\times d}$, characterizing the multi-modal cognitive state at the time, where $y$ and $z$ denote the lengths. 
Subsequently, we concatenate the initialized memory query to the rear of these hidden states to update the queried semantic information:
\vspace{-3pt}
\begin{equation}
    \mathbf{Q} = \mathcal{B}\left([\mathbf{H},\mathbf{Q}_{init}]\right)[-K:],
    \label{eq:query_builder}
\end{equation}
where we select the output of the last layer of the encoder  (see Eq.~({\color{cvprblue}{10}})), and take the last $K$ encoded vectors as the memory query $\mathbf{Q}\in \mathbb{R}^{K\times d}$ to hook latent memory. Furthermore, we employ a masked attention to exclusively enable attention propagation from the query to the hidden states $\mathbf{H}$, while suppressing attention in the reverse direction, \textit{i.e.}, from $\mathbf{H}$ to $\mathbf{Q}$ (see Eq.~({\color{cvprblue}{11}})). Here, both short- and long-term memory share the same query builder $\mathcal{B}$.

\vspace{2pt}
\noindent \textbf{Latent Memory Former.} Distinct from many existing paradigms~\cite{shen2025vlm,huang2025vision,yang2025machine}, we internalize the latent vision memory into lightweight formers, preserving the general abilities of base VLMs and ensuring the compatibility of our paradigm. We initialize two lightweight LoRA adapters, which are respectively designated as the short-term memory former $\mathcal{F}_{s}$ and long-term memory former $\mathcal{F}_{l}$, attached to the vision encoder and the final language model of the VLM, without directly tampering with the core parameters.
More precisely, we first append the generated memory query $\mathbf{Q}$ along with a set of learnable memory tokens after the corresponding target token sequence $\textbf{X}$. Then we process it by short-term or long-term memory former, which contextualizes and embeds the latent memory information:
\vspace{-3pt}
\begin{equation}
\mathbf{M}_{s/l}=\mathcal{F}_{s/l}\left(\left[\textbf{X},\mathbf{Q},\mathbf{M}_{init}\right]\right)\left[-N_{s/l}:\right],
\end{equation}
where short- and long-term latent vision memory $\mathbf{M}_{s/l}\in\mathbb{R}^{N_{s/l}\times d}$, while $N_s$ and $N_l$ are the predetermined lengths of memory tokens, which can be taken from $\left\{2,4,8,16,32\right\}$. For the short-term pathway, the resultant memory representation is concatenated with the visual token stream, and pass through the original projector to align it with the representation space of the language model.   
The two memory formers serve as dedicated memory carriers, exclusively storing visual evidences and semantic knowledge within themselves. When the policy model executes a memory invocation, the incoming memory query  triggers externalization of useful short- or long-term memory. These memories are seamlessly inserted into the token generation process alongside the invocation and end signals and barely interfere with the original generation, as specified in \cref{eq:autograssive}.

\subsection{Training Recipe}
\label{sec:training}
We design a two-stage training procedure based on GRPO~\cite{shao2024deepseekmath}, whose optimization objectives are to optimize the effective formation and invocation of latent memory. The first stage enhances the utility of memory, while the second stage maximizes the reward of each invocation, thereby accelerating the convergence of different components steadily. More detailed algorithms and implementations are present in Appendix~{\color{cvprblue}{7.2}} and {\color{cvprblue}{8.3}}.

\vspace{2pt}
\noindent \textbf{Stage I: Memory Formation Optimization.} In this stage, we update the query builder $\mathcal{B}$, and memory formers $\mathcal{F}_{s/l}$ while keeping the policy model $\mathcal{P}$ frozen. Initially, during the autoregressive generation process, we randomly invoke either short- or long-term  memory upon detecting the delimiter, thereby acquiring initial memory capabilities. Then, the scope of memory invocations is extended to the intervals between delimiters, this not only provides a richer trajectory of memory interactions but also enables memory invocation at arbitrary positions within the generation sequence. The core objective is to maximize the performance improvement relative to trajectory without memory integration $\Delta S(\tau)=S(\tau)-S(\tau_{base})$, thereby enhancing the quality of the memory formation (full function in Eq.~({\color{cvprblue}{14}})):
\vspace{-3pt}
\begin{equation}
        \max _{\mathcal{F}_{s/l}, \mathcal{B}} \mathbb{E}_{\tau \sim \mathcal{P}\left(\cdot \mid x,\mathbf{M}_{s/l}\right),\mathbf{M}_{s/l}\sim\mathcal{F}_{s/l}\left(\mathbf{Q}\right),\mathbf{Q}\sim\mathcal{B}\left(\mathbf{H}\right) }[\Delta S(\tau)].
        \label{eq:stage1_objective}
\end{equation}

\vspace{2pt}
\noindent \textbf{Stage II: Memory Invocation Optimization.} In this process, we update part parameters $\theta$ of the policy model $\mathcal{P}$, and keeps all the memory formation components frozen. At this stage, the policy model $\mathcal{P}$ is required to invoke memory efficiently and accurately, which entails two core requirements: selecting the correct memory type and avoiding invalid invocations. Thus, we add two penalties to the objective, which could be optimized by (full function in Eq.~({\color{cvprblue}{15}})):
\vspace{-3pt}
\begin{equation}
        \max _{\theta} \mathbb{E}_{\tau \sim \mathcal{P}\left(\cdot \mid x,\mathbf{M}_{s/l}\right)}[\Delta S(\tau)-\alpha(p_{type}+p_{neg})],
        \label{eq:stage2_objective}
\end{equation}
where $\alpha$ denotes the penalty intensity. The type penalty, $p_{\text{type}} = \max\left(0, S(\tau_{\text{rev}}) - S(\tau)\right)$, serves to penalize the erroneous selection of memory types, where $\tau_{\text{rev}}$ represents the invocation of an alternative memory type. In parallel, the negative penalty $p_{\text{neg}} = \max\left(0, \overline{S} - S(\tau)\right)$ is designed to penalize invocations with negative returns, aiming to enhance efficiency. Here, $\overline{S}$ denotes the mean of quantifiable scores across candidate trajectories.

%% file: sec/4_experiment.tex
\section{Experiments}

\begin{table*}[t]
\centering
\caption{Results on 12 benchmarks to evaluate visual understanding, reasoning and generation abilities. The \textbf{best} and \underline{second best} values are emphasized, and the average values are calculated for both specific capabilities and overall results. }
\vspace{-3mm}
\setlength{\tabcolsep}{0.9mm}
\resizebox{0.97\textwidth}{!}{
\begin{tabular}{l|cccccc|ccccc|cccc|c}
\toprule   
\multirow{2}{*}{\textbf{Method}} & \textbf{MM} & \textbf{MM} & \multirow{2}{*}{\textbf{MMT}} & \multirow{2}{*}{\textbf{BLINK}} & \textbf{Muir} & \multirow{2}{*}{\textbf{\textit{\color{black}{Avg.}}}} &  \multirow{2}{*}{\textbf{MMMU}} & \textbf{Logic} & \textbf{Math} & \textbf{MV} & \multirow{2}{*}{\textbf{\textit{\color{black}{Avg.}}}} & \textbf{Hall} & \textbf{Multi} & \multirow{2}{*}{\textbf{MMVU}} & \multirow{2}{*}{\textbf{\textit{\color{black}{Avg.}}}} & \multirow{2}{*}{\textbf{\textit{\color{black}{Avg.}}}} \\  
 & \textbf{Star} & \textbf{Vet} & & & \textbf{Bench} & & & \textbf{Vista} & \textbf{Vista} & \textbf{-Math}  & & \textbf{Bench} & \textbf{Trust} & & \\\midrule
Vanilla~\cite{bai2025qwen2} & 62.6 & 66.0 & 54.0 & 55.4 & 57.4 & \color{black}{\textit{ 59.3 }} & 56.0 & 43.5 & 67.8 & 18.9 & \color{black}{\textit{46.6}}& 52.3 & 64.8 & 55.4 & \color{black}{\textit{ 57.7 }} & \color{black}{\textit{54.5}}   \\ \midrule
SFT & 64.7 & 67.5 & 56.8 & 54.5 & 58.7 & \color{black}{\textit{ 60.3 }} & 57.7 & 46.1 & 69.5 & 22.8 & \color{black}{\textit{49.0}}& 53.6 & 67.0 & 59.1 & \color{black}{\textit{ 59.9 }} & \color{black}{\textit{56.5}}   \\

Visual-RFT~\cite{liu2025visual} & 65.6 & 70.5 & 59.1 & 58.0 & 62.9 & \color{black}{\textit{ 63.6 }} & 62.4 & 51.7 & 71.6 & 26.5 & \color{black}{\textit{53.0}}& 55.8 & 70.7 & 63.2 & \color{black}{\textit{ 63.2 }} & \color{black}{\textit{59.8}}   \\

VLM-R1~\cite{shen2025vlm} & 66.3 & 73.0 & 59.4 & 60.6 & 63.8 & \color{black}{\textit{ 64.6 }} & 63.4 & 53.0 & 75.9 & 34.6 & \color{black}{\textit{56.7}}& 54.2 & 69.9 & 61.7 & \color{black}{\textit{ 61.9 }} & \color{black}{\textit{61.3}}   \\
Vision-R1~\cite{huang2025vision} & \underline{67.1} & 71.7 & 60.2 & \underline{60.8} & \underline{64.0} & \color{black}{\textit{ \underline{65.0} }} & 63.2 & \underline{53.9} & 77.2 & 38.7 & \color{black}{\textit{ \underline{58.2} }}& 56.4 & 72.6 & 63.6 & \color{black}{\textit{ 64.2 }} & \color{black}{\textit{62.5}}   \\
PAPO~\cite{wang2025perception} & 64.2 & 69.8 & 57.9 & 53.3 & 56.7 & \color{black}{\textit{ 60.4 }} & 61.2 & 52.5 & 73.3 & 34.8 & \color{black}{\textit{55.5}}& 50.3 & 67.7 & 56.5 & \color{black}{\textit{ 58.2 }} & \color{black}{\textit{58.2}}   \\ \midrule
Sketchpad~\cite{hu2024visual} & 62.1 & 64.5 & 57.0 & 54.9 & 52.8 & \color{black}{\textit{ 58.3 }} & 57.9 & 47.4 & 68.4 & 24.6 & \color{black}{\textit{49.6}}& 52.1 & 66.2 & 57.2 & \color{black}{\textit{ 58.5 }} & \color{black}{\textit{55.4}}   \\
GRIT~\cite{fan2025grit} & 65.8 & 67.8 & 57.9 & 52.5 & 51.0 & \color{black}{\textit{ 59.0 }} & 59.4 & 51.6 & 68.1 & 22.4 & \color{black}{\textit{50.4}}& 53.7 & 67.3 & 60.1 & \color{black}{\textit{ 60.4 }} & \color{black}{\textit{56.5}}   \\
PixelReasoner~\cite{su2025pixel} & 65.3 & 67.1 & 58.7 & 56.8 & 60.5 & \color{black}{\textit{ 61.7 }} & 58.9 & 49.3 & 69.6 & 25.9 & \color{black}{\textit{50.9}}& 55.9 & 69.9 & 61.5 & \color{black}{\textit{ 62.4 }} & \color{black}{\textit{58.3}}   \\
DeepEyes~\cite{zheng2025deepeyes} & 66.4 & 70.5 & 60.3 & 60.4 & 63.0 & \color{black}{\textit{ 64.1 }} & 60.3 & 49.1 & 70.8 & 31.5 & \color{black}{\textit{52.9}}& 57.4 & 72.6 & 64.6 & \color{black}{\textit{ \underline{64.9} }} & \color{black}{\textit{60.5}}   \\
OpenThinkImg~\cite{su2025openthinkimg} & 66.0 & 71.6 & 60.8 & 59.2 & 61.7 & \color{black}{\textit{ 63.9 }} & 61.4 & 52.8 & 73.0 & 28.0 & \color{black}{\textit{53.8}}& 54.9 & 74.0 & 64.3 & \color{black}{\textit{ 64.4 }} & \color{black}{\textit{60.6}}   \\ \midrule
Scaffold~\cite{lei2024scaffolding} & 63.9 & 67.0 & 58.5 & 52.5 & 52.9 & \color{black}{\textit{ 59.0 }} & 58.1 & 51.0 & 64.7 & 21.0 & \color{black}{\textit{48.7}}& 54.8 & 68.5 & 60.6 & \color{black}{\textit{ 61.3 }} & \color{black}{\textit{56.1}}   \\
ICoT~\cite{gao2025interleaved} & 65.6 & 67.9 & 60.5 & 54.3 & 57.0 & \color{black}{\textit{ 61.1 }} & 58.6 & 49.8 & 76.7 & 30.8 & \color{black}{\textit{54.0}}& 57.0 & 69.1 & 62.0 & \color{black}{\textit{ 62.7 }} & \color{black}{\textit{59.1}}   \\
MINT-CoT~\cite{chen2025mint} & 66.2 & 69.5 & 57.3 & 55.4 & 58.9 & \color{black}{\textit{ 61.5 }} & 57.7 & 51.5 & \underline{77.4} & 39.2 & \color{black}{\textit{56.5}}& 56.7 & 71.4 & 60.8 & \color{black}{\textit{ 63.0 }} & \color{black}{\textit{60.2}}   \\
VPT~\cite{yu2025introducing} & 64.2 & 70.8 & 59.0 & 58.6 & 63.5 & \color{black}{\textit{ 63.2 }} & 59.1 & 53.0 & 72.3 & 34.7 & \color{black}{\textit{54.8}}& 52.3 & 64.7 & 61.4 & \color{black}{\textit{ 59.5 }} & \color{black}{\textit{59.5}}   \\ \midrule
Mirage~\cite{yang2025machine} & 64.5 & 71.8 & 56.1 & 56.3 & 59.0 & \color{black}{\textit{ 61.5 }} & 59.4 & 50.6 & 70.3 & 35.4 & \color{black}{\textit{53.9}}& 50.9 & 66.1 & 60.3 & \color{black}{\textit{ 59.1 }} & \color{black}{\textit{58.4}}   \\
\rowcolor{gray!20} \textbf{VisMem (Ours)}   &  \textbf{68.9}  &  \textbf{75.1}  &   \textbf{62.5}  &   \textbf{64.5}  &   \textbf{69.8}   & \color{black}{\textit{ \textbf{68.2} }} &   \textbf{63.9}  &   \textbf{55.7}   &  \textbf{79.8}  &   \textbf{41.4}   & \color{black}{\textit{60.2}} &   \textbf{59.6}  &   \textbf{77.0}   &   \textbf{68.2}  & \color{black}{\textit{ \textbf{68.3} }} & \color{black}{\textit{65.5}}   \\  \bottomrule
\end{tabular}}
\label{tab:comparison}
\vspace{-12pt}
\end{table*}

\subsection{Settings}

\noindent\textbf{Benchmarks.}
We select 12 benchmarks to comprehensively evaluate three main abilities of VLMs, \textit{i.e.}, understanding, reasoning and generation~\cite{li2024survey}. These benchmarks include:
\textbf{(1) understanding:} MMStar~\cite{chen2024we}, MMVet~\cite{yu2024mmvet}, MMT~\cite{ying2024mmt}, BLINK~\cite{fu2024blink}, MuirBench~\cite{wang2025muirbench};
\textbf{(2) reasoning:} MMMU~\cite{yue2024mmmu}, LogicVista~\cite{xiao2024logicvista}, MathVista~\cite{lu2024mathvista}, MV-Math~\cite{wang2025mv};
\textbf{(3) generation:} HallBench~\cite{guan2024hallusionbench}, MultiTrust~\cite{zhang2024multitrust}, MMVU~\cite{liu2025unveiling}. Details are in Appendix~{\color{cvprblue}{8.2}}.

\vspace{2pt}
\noindent\textbf{Baselines.} We compare our VisMem against 15 baselines, falling into four categories: 
\textbf{(a) direct training methods}: SFT, Visual-RFT~\cite{liu2025visual}, VLM-R1~\cite{shen2025vlm}, Vision-R1~\cite{huang2025vision} and PAPO~\cite{wang2025perception}; 
\textbf{(b) image-level methods}: GRIT~\cite{fan2025grit}, Sketchpad~\cite{hu2024visual}, MVoT~\cite{li2025imagine}, OpenThinkImg~\cite{su2025openthinkimg} and DeepEyes~\cite{zheng2025deepeyes};
\textbf{(c) token-level methods}: Scaffold~\cite{lei2024scaffolding}, MINT-CoT~\cite{chen2025mint}, ICoT~\cite{gao2025interleaved}, and VPT~\cite{yu2025introducing}; 
\textbf{(d) latent space methods}: Mirage~\cite{yang2025machine}. Details are in Appendix~{\color{cvprblue}{8.3}}.

\vspace{2pt}
\noindent\textbf{Implementation Details.}
All experiments (except for \cref{tab:base_models}) are implemented on Qwen2.5-VL-7B~\cite{bai2025qwen2} based on 8 NVIDIA H200 141G GPUs. The length of memory query $K$ is set to 8, and the lengths of short-term $N_s$ and long-term latent vision memory $N_l$ are 8 and 16, respectively. More implementation details are listed in Appendix~{\color{cvprblue}{8.4}}.

\subsection{Main Results}
The main experimental results demonstrate that our proposed memory system VisMem unlocks the untapped potentials with three key \textit{\textbf{enh}}ancements: \textit{\textbf{[Enh.1]}} advanced visual capabilities, \textit{\textbf{[Enh.2]}} cross-domain generalization, \textit{\textbf{[Enh.3]}} catastrophic forgetting alleviation.

\begin{table*}[t]
\centering
\caption{Results on nine base models with various sizes and sources, including Qwen2.5-VL-3B/7B/32B~\cite{bai2025qwen2}, LLaVA-OV-1.5-4B/8B~\cite{an2025llava}, InternVL-3.5-4B/8B/14B/38B~\cite{wang2025internvl3}. {\color{mygreen}{$\uparrow$}} indicates the performance enhancement compared with the base model.}
\vspace{-3mm}
\setlength{\tabcolsep}{0.9mm}
\resizebox{0.99\textwidth}{!}{
\begin{tabular}{l|lllll|llll|lll}
\toprule   
\multirow{2}{*}{\textbf{Base Model}} & \textbf{MM} & \textbf{MM} & \multirow{2}{*}{\textbf{MMT}} & \multirow{2}{*}{\textbf{BLINK}} & \textbf{Muir} &  \multirow{2}{*}{\textbf{MMMU}} & \textbf{Logic} & \textbf{Math} & \textbf{MV} & \textbf{Hall} & \textbf{Multi} & \multirow{2}{*}{\textbf{MMVU}} \\  
 & \textbf{Star} & \textbf{Vet} & & & \textbf{Bench} & & \textbf{Vista} & \textbf{Vista} & \textbf{-Math} & \textbf{Bench} & \textbf{Trust}  \\\midrule
 
Qwen2.5-VL-3B~\cite{bai2025qwen2} & 52.9 & 61.5 & 49.8 & 46.0 & 46.1 & 52.6 & 39.7 & 61.0 & 13.2 & 46.3 & 56.9 & 48.4 \\
\rowcolor{gray!20} \textbf{+ VisMem (Ours)}  &  61.0 {\color{mygreen}{\footnotesize{$\uparrow$}8.1}} &  72.5 {\color{mygreen}{\footnotesize{$\uparrow$}11.0}} &  59.3 {\color{mygreen}{\footnotesize{$\uparrow$}9.5}} &  58.6 {\color{mygreen}{\footnotesize{$\uparrow$}12.6}} &  64.4 {\color{mygreen}{\footnotesize{$\uparrow$}18.3}} &  61.9 {\color{mygreen}{\footnotesize{$\uparrow$}9.3}} &  53.1 {\color{mygreen}{\footnotesize{$\uparrow$}13.4}} &  70.4 {\color{mygreen}{\footnotesize{$\uparrow$}9.4}} &  31.7 {\color{mygreen}{\footnotesize{$\uparrow$}18.5}} &  58.0 {\color{mygreen}{\footnotesize{$\uparrow$}11.7}} &  70.3 {\color{mygreen}{\footnotesize{$\uparrow$}13.4}} &  60.6 {\color{mygreen}{\footnotesize{$\uparrow$}12.2}} \\
Qwen2.5-VL-7B~\cite{bai2025qwen2}  & 62.6 & 66.0 & 54.0 & 55.4 & 57.4 & 56.0 & 43.5 & 67.8 & 18.9 & 52.3 & 64.8 & 55.4 \\
\rowcolor{gray!20} \textbf{+ VisMem (Ours)}  &  68.9 {\color{mygreen}{\footnotesize{$\uparrow$}6.3}} &  75.1 {\color{mygreen}{\footnotesize{$\uparrow$}9.1}} &  62.5 {\color{mygreen}{\footnotesize{$\uparrow$}8.5}} &  64.5 {\color{mygreen}{\footnotesize{$\uparrow$}9.1}} &  69.8 {\color{mygreen}{\footnotesize{$\uparrow$}11.4}} &  63.9 {\color{mygreen}{\footnotesize{$\uparrow$}7.9}} &  55.7 {\color{mygreen}{\footnotesize{$\uparrow$}12.2}} &  79.8 {\color{mygreen}{\footnotesize{$\uparrow$}12.0}} &  41.4 {\color{mygreen}{\footnotesize{$\uparrow$}22.5}} &  59.6 {\color{mygreen}{\footnotesize{$\uparrow$}7.3}} &  77.0 {\color{mygreen}{\footnotesize{$\uparrow$}12.2}} &  68.2 {\color{mygreen}{\footnotesize{$\uparrow$}12.8}} \\
Qwen2.5-VL-32B~\cite{bai2025qwen2} & 67.1 & 68.7 & 64.7 & 59.9 & 63.5 & 70.6 & 47.9 & 72.7 & 29.0 & 53.6 & 64.5 & 55.5\\
\rowcolor{gray!20} \textbf{+ VisMem (Ours)} &  73.9 {\color{mygreen}{\footnotesize{$\uparrow$}6.8}} &  77.9 {\color{mygreen}{\footnotesize{$\uparrow$}9.2}} &  72.0 {\color{mygreen}{\footnotesize{$\uparrow$}7.3}} &  68.6 {\color{mygreen}{\footnotesize{$\uparrow$}8.7}} &  73.3 {\color{mygreen}{\footnotesize{$\uparrow$}9.8}} &  75.9 {\color{mygreen}{\footnotesize{$\uparrow$}5.3}} &  63.5 {\color{mygreen}{\footnotesize{$\uparrow$}15.6}} & 83.5  {\color{mygreen}{\footnotesize{$\uparrow$}10.8}} &  54.9 {\color{mygreen}{\footnotesize{$\uparrow$}25.9}} &  60.2 {\color{mygreen}{\footnotesize{$\uparrow$}6.6}} &  77.7 {\color{mygreen}{\footnotesize{$\uparrow$}13.2}} &  68.4 {\color{mygreen}{\footnotesize{$\uparrow$}12.9}} \\ \midrule
LLaVA-OV-1.5-4B~\cite{an2025llava} & 62.5 & 60.4 & 54.4 & 38.2 & 42.6 & 49.4 & 39.3 & 66.5 & 11.0 & 41.8 & 47.5 & 44.2 \\
\rowcolor{gray!20} \textbf{+ VisMem (Ours)}  &  69.0 {\color{mygreen}{\footnotesize{$\uparrow$}6.5}} &  70.1 {\color{mygreen}{\footnotesize{$\uparrow$}9.7}} &  62.7 {\color{mygreen}{\footnotesize{$\uparrow$}8.3}} &  56.9 {\color{mygreen}{\footnotesize{$\uparrow$}18.7}} &  59.6 {\color{mygreen}{\footnotesize{$\uparrow$}17.0}} &  59.7 {\color{mygreen}{\footnotesize{$\uparrow$}10.3}} &  53.7 {\color{mygreen}{\footnotesize{$\uparrow$}14.4}} &  79.0 {\color{mygreen}{\footnotesize{$\uparrow$}12.5}} &  27.2 {\color{mygreen}{\footnotesize{$\uparrow$}16.2}} &  52.8 {\color{mygreen}{\footnotesize{$\uparrow$}11.0}} &  66.4 {\color{mygreen}{\footnotesize{$\uparrow$}18.9}} &  61.9 {\color{mygreen}{\footnotesize{$\uparrow$}17.7}} \\
LLaVA-OV-1.5-8B~\cite{an2025llava} & 65.3 & 67.1 & 57.8 & 49.8 & 50.5 & 55.3 & 46.5 & 68.3 & 15.7 & 50.1 & 54.7 & 50.6 \\
\rowcolor{gray!20} \textbf{+ VisMem (Ours)}  &  70.8 {\color{mygreen}{\footnotesize{$\uparrow$}5.5}} &  75.7 {\color{mygreen}{\footnotesize{$\uparrow$}8.6}} &  64.7 {\color{mygreen}{\footnotesize{$\uparrow$}6.9}} &  61.0 {\color{mygreen}{\footnotesize{$\uparrow$}11.2}} &  62.6 {\color{mygreen}{\footnotesize{$\uparrow$}12.1}} &  63.0 {\color{mygreen}{\footnotesize{$\uparrow$}7.7}} &  59.5 {\color{mygreen}{\footnotesize{$\uparrow$}13.0}} &  80.0 {\color{mygreen}{\footnotesize{$\uparrow$}11.7}} &  34.5 {\color{mygreen}{\footnotesize{$\uparrow$}18.8}} &  55.5 {\color{mygreen}{\footnotesize{$\uparrow$}5.4}} &  69.4 {\color{mygreen}{\footnotesize{$\uparrow$}14.7}} &  67.0 {\color{mygreen}{\footnotesize{$\uparrow$}16.4}} \\ \midrule 
InternVL-3.5-4B~\cite{wang2025internvl3} & 62.8 & 73.1 & 62.7 & 57.1 & 52.8 & 59.9 & 53.2 & 76.3 & 17.5 & 43.0 & 56.2 & 44.7 \\ 
\rowcolor{gray!20} \textbf{+ VisMem (Ours)}  &  70.2 {\color{mygreen}{\footnotesize{$\uparrow$}7.4}} &  80.3 {\color{mygreen}{\footnotesize{$\uparrow$}7.2}} &  69.0 {\color{mygreen}{\footnotesize{$\uparrow$}6.3}} &  65.2 {\color{mygreen}{\footnotesize{$\uparrow$}8.1}} &  63.9 {\color{mygreen}{\footnotesize{$\uparrow$}11.1}} &  68.5 {\color{mygreen}{\footnotesize{$\uparrow$}8.6}} &  64.6 {\color{mygreen}{\footnotesize{$\uparrow$}11.4}} &  82.5 {\color{mygreen}{\footnotesize{$\uparrow$}6.2}} &  30.8 {\color{mygreen}{\footnotesize{$\uparrow$}13.3}} &  52.4 {\color{mygreen}{\footnotesize{$\uparrow$}9.4}} &  70.4 {\color{mygreen}{\footnotesize{$\uparrow$}14.2}} &  61.9 {\color{mygreen}{\footnotesize{$\uparrow$}17.2}} \\
InternVL-3.5-8B~\cite{wang2025internvl3} & 67.0 & 80.1 & 64.6 & 58.4 & 55.7 & 68.6 & 54.8 & 77.5 & 27.1 & 54.5 & 65.9 & 52.3 \\ 
\rowcolor{gray!20} \textbf{+ VisMem (Ours)} & 71.8 {\color{mygreen}{\footnotesize{$\uparrow$}4.8}} &  85.4 {\color{mygreen}{\footnotesize{$\uparrow$}5.3}} &  69.5 {\color{mygreen}{\footnotesize{$\uparrow$}4.9}} &  66.1 {\color{mygreen}{\footnotesize{$\uparrow$}7.7}} &  65.3 {\color{mygreen}{\footnotesize{$\uparrow$}9.6}} &  73.3 {\color{mygreen}{\footnotesize{$\uparrow$}4.7}} &  64.7 {\color{mygreen}{\footnotesize{$\uparrow$}9.9}} &  82.9 {\color{mygreen}{\footnotesize{$\uparrow$}5.4}} &  44.7 {\color{mygreen}{\footnotesize{$\uparrow$}17.6}} &  60.9 {\color{mygreen}{\footnotesize{$\uparrow$}6.4}} &  78.5 {\color{mygreen}{\footnotesize{$\uparrow$}12.6}} &  68.3 {\color{mygreen}{\footnotesize{$\uparrow$}16.0}} \\  
InternVL-3.5-14B~\cite{wang2025internvl3} & 67.3 & 79.0 & 66.1 & 56.9 & 57.7 & 68.8 & 55.9 & 79.1 & 29.4 & 54.1 & 69.6 & 54.8 \\ 
\rowcolor{gray!20} \textbf{+ VisMem (Ours)}  &  72.4 {\color{mygreen}{\footnotesize{$\uparrow$}5.1}} &  85.7 {\color{mygreen}{\footnotesize{$\uparrow$}6.7}} &  70.6 {\color{mygreen}{\footnotesize{$\uparrow$}4.5}} &  66.1 {\color{mygreen}{\footnotesize{$\uparrow$}9.2}} &  67.0 {\color{mygreen}{\footnotesize{$\uparrow$}9.3}} &  73.8 {\color{mygreen}{\footnotesize{$\uparrow$}5.0}} &  65.5 {\color{mygreen}{\footnotesize{$\uparrow$}9.6}} &  85.1 {\color{mygreen}{\footnotesize{$\uparrow$}6.0}} &  46.6 {\color{mygreen}{\footnotesize{$\uparrow$}17.2}} &  60.5 {\color{mygreen}{\footnotesize{$\uparrow$}6.4}} &  77.8 {\color{mygreen}{\footnotesize{$\uparrow$}8.2}} &  68.3 {\color{mygreen}{\footnotesize{$\uparrow$}13.5}} \\ 
InternVL-3.5-38B~\cite{wang2025internvl3} & 72.2 & 79.7 & 70.5 & 60.3 & 64.0 & 72.1 & 61.1 & 80.2 & 35.7 & 60.2 & 71.5 & 58.0 \\ 
\rowcolor{gray!20} \textbf{+ VisMem (Ours)}  &  75.1 {\color{mygreen}{\footnotesize{$\uparrow$}2.9}} &  86.4 {\color{mygreen}{\footnotesize{$\uparrow$}6.7}} &  73.3 {\color{mygreen}{\footnotesize{$\uparrow$}2.8}} &  67.5 {\color{mygreen}{\footnotesize{$\uparrow$}7.2}} &  69.9 {\color{mygreen}{\footnotesize{$\uparrow$}5.9}} &  75.8 {\color{mygreen}{\footnotesize{$\uparrow$}3.7}} &  68.7 {\color{mygreen}{\footnotesize{$\uparrow$}7.6}} &  85.4 {\color{mygreen}{\footnotesize{$\uparrow$}5.2}} &  56.9 {\color{mygreen}{\footnotesize{$\uparrow$}21.2}} &  65.8 {\color{mygreen}{\footnotesize{$\uparrow$}5.6}} &  79.0 {\color{mygreen}{\footnotesize{$\uparrow$}7.5}} &  69.9 {\color{mygreen}{\footnotesize{$\uparrow$}11.9}} \\ \bottomrule
\end{tabular}}
\label{tab:base_models}
\vspace{-12pt}
\end{table*}

\vspace{2pt}
\noindent \textbf{\textit{[Enh.1]} VisMem enables advanced and comprehensive visual capabilities.} As presented in \cref{tab:comparison}, our proposed method demonstrates distinct superiority over other baseline models. Compared with the vanilla model, VisMem achieves a notable average improvement of \textbf{11.0\%} across all benchmarks.
When compared with the top three baselines (\textit{i.e.}, Vision-R1~\cite{huang2025vision}, VLM-R1~\cite{shen2025vlm}, and OpenThinkImg~\cite{su2025openthinkimg}), our method still maintains improvements of 3.0\%, 4.2\%, and 4.9\%, respectively. 
Furthermore, it consistently enhances performance across the three core domains of visual tasks, namely, understanding, reasoning, and generation.  Our latent vision memory mechanism yields comprehensive enhancements in visual capabilities, with specific gains of +8.9\% in visual understanding, +14.4\% in reasoning, and +10.6\% in generation, relative to the vanilla model.
It is also noteworthy that direct RL-based methods (\textit{e.g.}, VLM-R1~\cite{shen2025vlm} and Vision-R1~\cite{huang2025vision}) also achieve relatively better performance than most other paradigms. However, this approach of directly modifying parameters relies on incremental parameter updates, which may lead to the overwriting of prior general knowledge and result in catastrophic forgetting.


As illustrated in Tab.~{\color{cvprblue}{5}} and {\color{cvprblue}{6}}, we conduct additional evaluations on selected subsets of MuirBench~\cite{wang2025muirbench} and LogicVista~\cite{xiao2024logicvista}. Endowed with short- and long-term vision memory, our VisMem outperforms all baseline methods by a substantial margin in tasks demanding fine-grained visual evidence, such as counting (+7.0\%), visual retrieval (+9.4\%), and grounding (13.1\%), while also yielding notable improvements in visual reasoning tasks, including inductive (+5.7\%) and deductive (+7.1\%) learning.


\begin{figure}[t]
    \centering
    \includegraphics[width=0.78\linewidth]{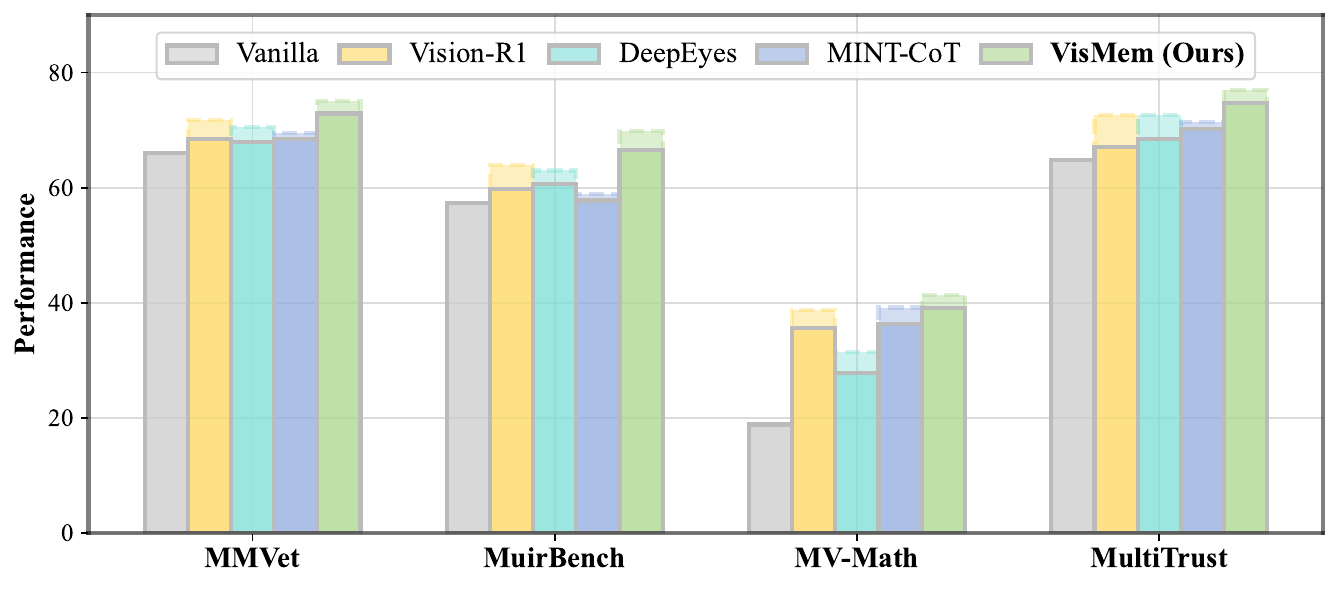}
    \vspace{-3mm}
    \caption{Results of the cross-domain generalization study. Models are only trained on Visual CoT~\cite{shao2024visual} and Mulberry~\cite{yao2024mulberry}. Dashed bar indicates the results with full training data.}
    \label{fig:cross_domain}
    \vspace{-16pt}
\end{figure}

\vspace{2pt}
\noindent \textbf{\textit{[Enh.2]} VisMem showcases great cross-domain generalization.} To evaluate the cross-domain generalization capability of our model, specifically whether its stored latent visual memory can transfer across diverse unseen tasks, we exclusively train our VisMem and comparative baseline models on two datasets: Visual CoT~\cite{shao2024visual} and Mulberry~\cite{yao2024mulberry}, then subsequently assess their performance on four unseen target benchmarks. As demonstrated in \cref{fig:cross_domain}, {\color{cvprblue}{7}}, and Tab.~{\color{cvprblue}{7}}, VisMem not only consistently achieves significant performance gains on out-of-domain tasks (+6.9\% on MMVet~\cite{yu2024mmvet}, +9.1\% on MuirBench~\cite{wang2025muirbench}, +20.2\% on MV-Math~\cite{wang2025mv}, and +9.9\% on MultiTrust~\cite{zhang2024multitrust}), but also maintains leading performance relative to all baselines. Notably, our method outperforms the second-ranked model by a substantial margin of 2.7–6.8\% across all four benchmarks, while narrowing the performance gap relative to results obtained with full training data. This observation underscores its robust cross-domain knowledge transfer capability.

\begin{figure}[t]
\includegraphics[width=0.78\linewidth]{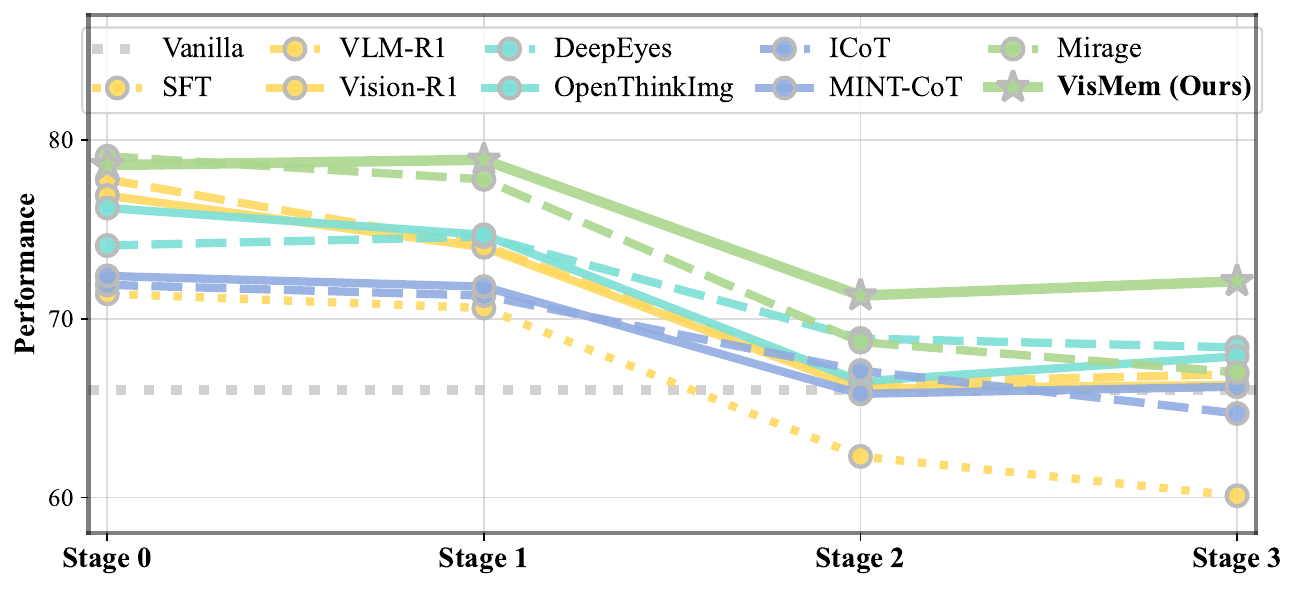}
\centering
    \vspace{-3mm}
    \caption{Results of four-stage continual learning on MMVet~\cite{yu2024mmvet}. Stage 0 only includes itself, while stage 1, 2, 3 sequentially train models on different additional training data combinations.}
    \label{fig:forget}
    \vspace{-16pt}
\end{figure}

\vspace{2pt}
\noindent \textbf{\textit{[Enh.3]} VisMem alleviates catastrophic forgetting.} As illustrated in \cref{fig:forget}, {\color{cvprblue}{8}}, and Tab.~{\color{cvprblue}{8}}, we conduct sequential training of the models across four stages, with performance assessed on MMVet~\cite{yu2024mmvet} after each stage. At stage 0, the model was trained exclusively on the base task, and in subsequent stages, we incrementally incorporated selected benchmarks into the training process.
From the continual learning results, our VisMem demonstrates significantly stronger knowledge retention capabilities. Although direct training paradigms yield relatively excellent overall performance in offline learning tasks with once-off training, they suffer from severe catastrophic forgetting. For instance, SFT exhibits over 10\% performance degradation throughout the training process, the highest among all baselines. Additionally, at stage 0, VLM-R1~\cite{shen2025vlm} and Vision-R1~\cite{liu2025visual} achieve performance improvements of 11.8\% and 10.9\% respectively compared to the vanilla model, however, these improvements are retained by less than 0.5\% at stage 4.
In contrast, our method effectively mitigates catastrophic forgetting, exhibiting the smallest performance gap relative to original full-data training among all baselines. It is further worth noting that our latent vision memory enhances performance at stages 1 and 3 without any degradation, reflecting superior cross-task generalization.

\subsection{Additional Analyses}
Through additional analyses, we derive three key research \textit{\textbf{obs}}ervations pertaining to VisMem: \textit{\textbf{[Obs.1]}} compatibility across base models, \textit{\textbf{[Obs.2]}} dynamic and adaptive memory invocation, \textit{\textbf{[Obs.3]}} relatively low inference latency.

\vspace{2pt}
\noindent \textbf{\textit{[Obs.1]} VisMem is robustly compatible across various base models.} As detailed in \cref{tab:base_models} and Fig.~{\color{cvprblue}{11}}, to evaluate the generalizability of our approach across diverse base models, we assess nine widely used base models, encompassing Qwen2.5-VL-3B/32B~\cite{bai2025qwen2}, LLaVA-OV-1.5-4B/8B~\cite{an2025llava}, InternVL-3.5-4B/8B/14B/38B~\cite{wang2025internvl3}, with parameter scales ranging from 3B to 38B. The results indicate that our latent vision memory paradigm exhibits strong compatibility across various models, yielding significant performance improvements across most visual tasks.

\begin{figure}[t]
    \centering 
    \begin{subfigure}[c]{0.235\textwidth}
        \centering
        \includegraphics[width=\textwidth]{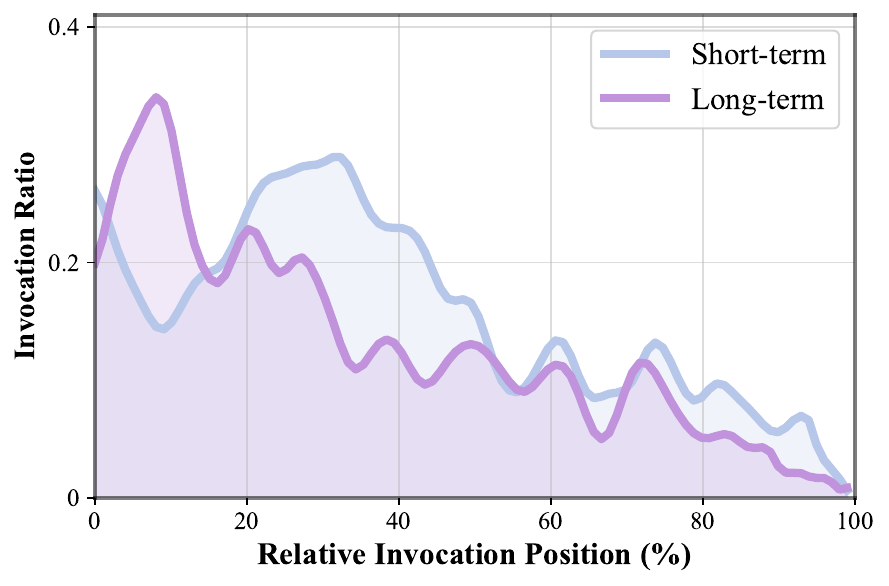} 
        \caption{MMVet~\cite{yu2024mmvet}} 
    \end{subfigure}
    \hfill  
    \begin{subfigure}[c]{0.235\textwidth}
        \centering
        \includegraphics[width=\textwidth]{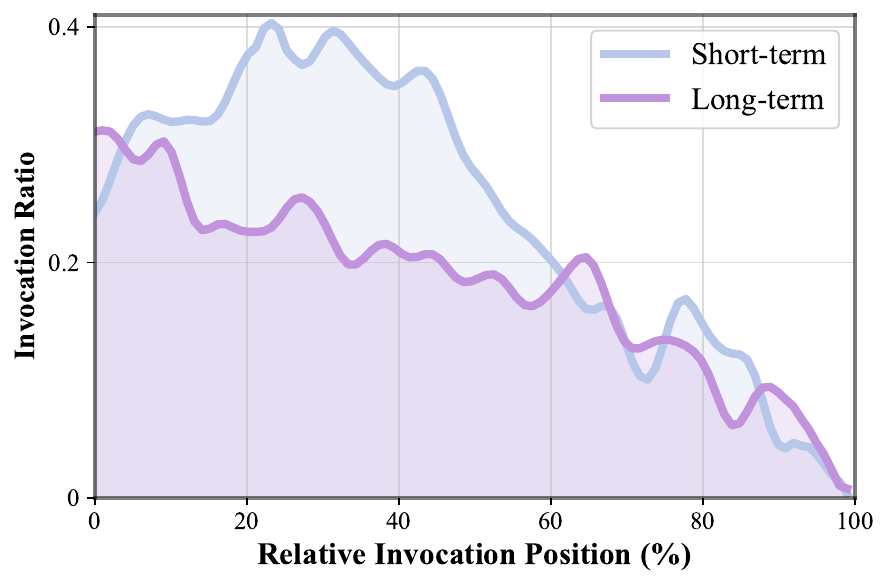}
        \caption{MuirBench~\cite{wang2025muirbench}}  
    \end{subfigure}

    \begin{subfigure}[c]{0.235\textwidth}
        \centering
        \includegraphics[width=\textwidth]{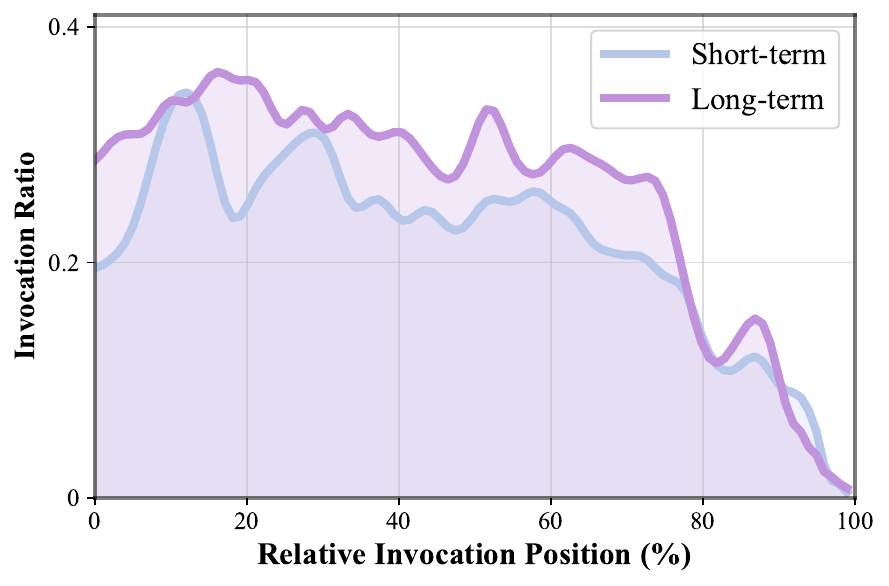}
        \caption{MV-Math~\cite{wang2025mv}}  
    \end{subfigure}
    \hfill
    \begin{subfigure}[c]{0.235\textwidth}
        \centering
        \includegraphics[width=\textwidth]{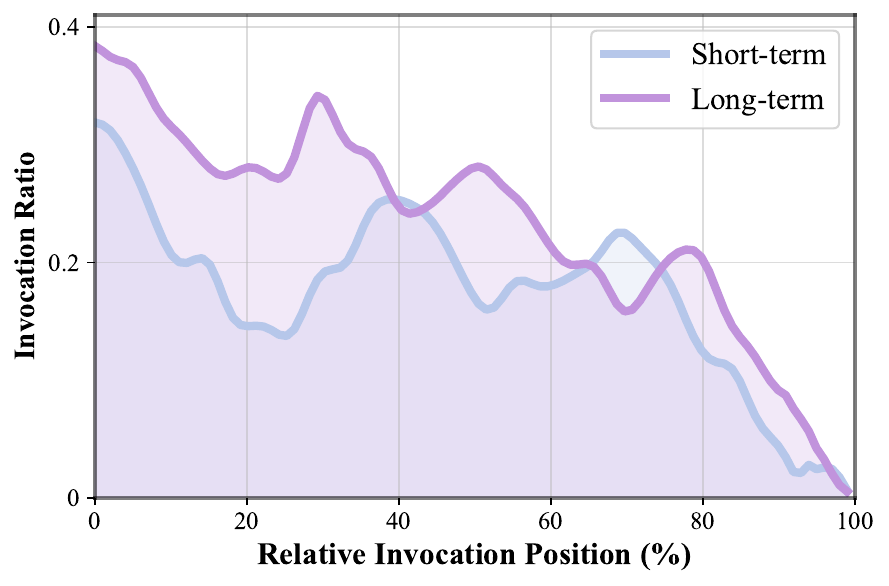}
        \caption{MultiTrust~\cite{zhang2024multitrust}} 
    \end{subfigure}
    \vspace{-3mm}
    \caption{Results of memory invocation ratio and invocation relative position across four benchmarks.} 
    \label{fig:memory_invocation} 
    \vspace{-16pt}
\end{figure}

\vspace{2pt}
\noindent \textbf{\textit{[Obs.2]} The memory invocations are dynamic and self-adaptive.} To elaborate on the effectiveness of our dual latent memory system, we characterize the properties of the short- and long-term memories it forms. As illustrated in \cref{fig:memory_invocation}, we first analyze the type-specific invocation ratios and their relative positions within the output sequence across four benchmarks.
In summary, invocation ratios are self-adaptive across tasks, while both memory types exhibit a dynamic downward trend in invocation frequency throughout the output sequence. Task-specific comparisons in Fig.~{\color{cvprblue}{9}} further reveal that short-term latent memories are invoked more frequently to retrieve fine-grained details during visual information acquisition and understanding, particularly in multi-image scenarios, such as MuirBench~\cite{wang2025muirbench}. Conversely, long-term latent vision memories play a more critical role in reasoning, \textit{e.g.}, in MV-Math~\cite{wang2025mv}, by providing abstract semantic knowledge relevant to the current task.
Furthermore, Tab.~{\color{cvprblue}{5}} and {\color{cvprblue}{6}}, which detail the sub-task performance of MuirBench~\cite{wang2025muirbench} and LogicVista~\cite{xiao2024logicvista} respectively, further illustrate that short-term and long-term latent visual memories are complementary. Their dynamic invocation yields superior performance compared to relying on a single memory type or the absence of vision memory.

\begin{figure}[t]
    \centering 
    \begin{subfigure}[c]{0.235\textwidth}
        \centering
        \includegraphics[width=\textwidth]{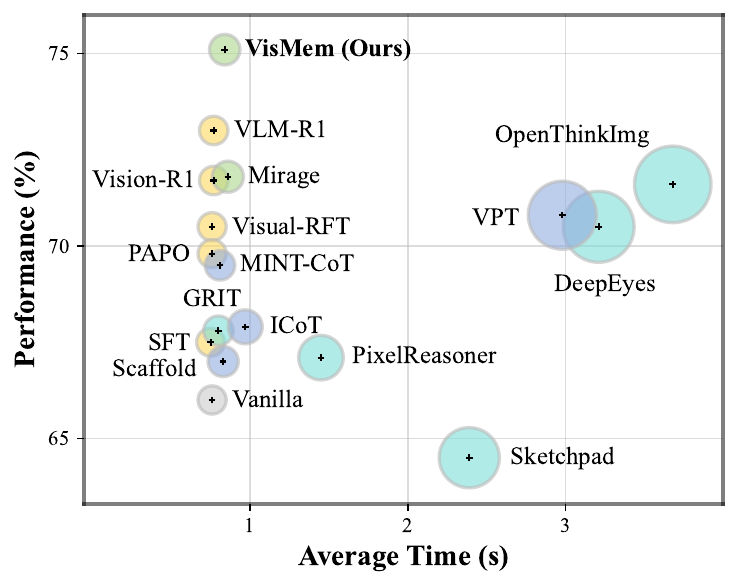} 
        \caption{MMVet~\cite{yu2024mmvet}} 
    \end{subfigure}
    \hfill  
    \begin{subfigure}[c]{0.235\textwidth}
        \centering
        \includegraphics[width=\textwidth]{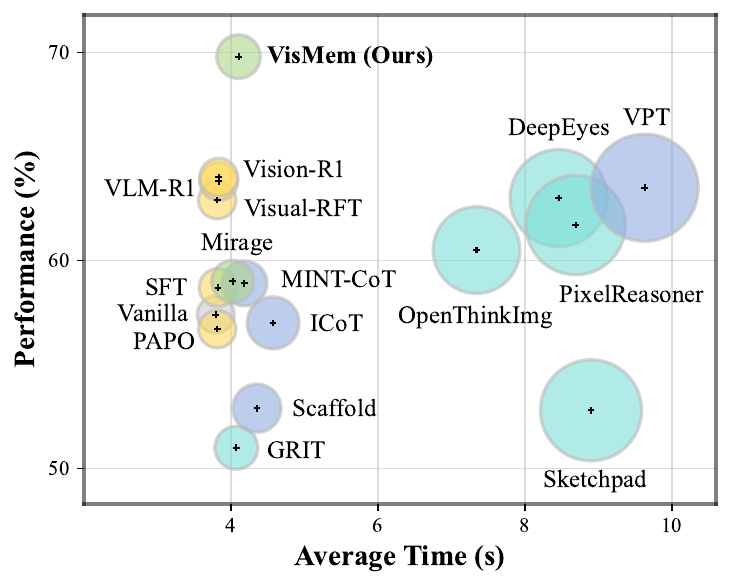}
        \caption{MuirBench~\cite{wang2025muirbench}}  
    \end{subfigure}

    \begin{subfigure}[c]{0.235\textwidth}
        \centering
        \includegraphics[width=\textwidth]{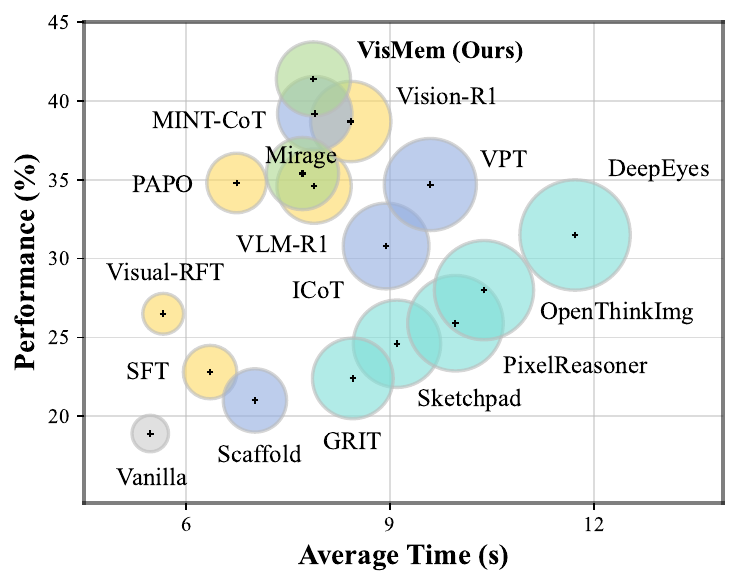}
        \caption{MV-Math~\cite{wang2025mv}}  
    \end{subfigure}
    \hfill
    \begin{subfigure}[c]{0.235\textwidth}
        \centering
        \includegraphics[width=\textwidth]{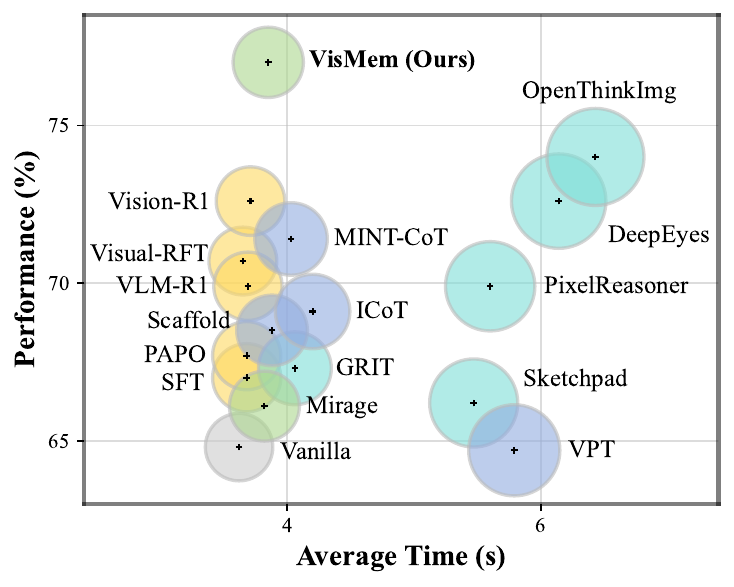}
        \caption{MultiTrust~\cite{zhang2024multitrust}} 
    \end{subfigure}
    \vspace{-3mm}
    \caption{Results of average inference time and performance across four benchmarks. The size is proportional to its y-value.} 
    \label{fig:efficiency} 
    \vspace{-6pt}
\end{figure}

\vspace{2pt}
\noindent \textbf{\textit{[Obs.3]} VisMem incurs minimal inference latency while yielding substantial performance gains.} As showcased in \cref{fig:efficiency} and Tab.~{\color{cvprblue}{12}}, we compare the average inference time and task performance on four benchmarks to quantify the efficiency-performance trade-off of our method. Our VisMem, by harnessing the capabilities of dual vision memory, attains the best performance while incurring insignificant inference latency.
Notably, image-level paradigms significantly elevate inference latency, particularly for tasks involving long thinking paths. In contrast, our VisMem exhibits remarkable effectiveness while maintaining average inference latency comparable to that of direct training optimization and token-level methods.

\vspace{2pt}
\noindent \textbf{Ablation Study and Sensitivity Analysis.} As reported in \cref{tab:ablation}, we conduct ablative studies on the memory invocation and dual memory formation. The results reveal that both short-term and long-term memory components contribute to performance across diverse visual tasks, while their complementarity synergistically drives the optimal performance. Additionally, as detailed in Tab.~{\color{cvprblue}{9}}, our design achieves a favorable balance between effectiveness and efficiency, with accurate and non-redundant memory invocation.
As shown in Fig.~{\color{cvprblue}{10}} and Tab.~{\color{cvprblue}{10}}, {\color{cvprblue}{11}}, we conduct sensitivity analyses of the sequence lengths of the memory query $K$, short-term $N_s$ and long-term $N_l$ latent memory tokens. As observed, performance generally improves with increasing sequence lengths within a reasonable range. Notably, our selected hyper-parameters achieve a favorable balance between performance and computational efficiency.

\begin{table}[t]
\centering
\caption{Ablations of latent vision memory invocation and dual latent vision memory formation.}
\vspace{-3mm}
\setlength{\tabcolsep}{0.9mm}
\resizebox{0.46\textwidth}{!}{
\begin{tabular}{l|cccc}
\toprule   
\multirow{1}{*}{\textbf{Ablation}} & \multicolumn{1}{c}{\textbf{MMVet}} & \multicolumn{1}{c}{\textbf{MuirBench}} & \multicolumn{1}{c}{\textbf{MV-Math}} & \multicolumn{1}{c}{\textbf{MultiTrust}} \\ \midrule
Vanilla & 66.0 & 57.4 & 18.9 & 64.8 \\ \midrule
Random Invocation (25\%)  &  69.2  &  59.4  &  29.8  &  69.4 \\
Random Invocation (50\%)  &  71.9  &  63.2  &  26.1  &  68.5 \\
Random Invocation (75\%)  &  \underline{73.6}  &  62.7  &  21.9  &  63.7 \\
Full Invocation (100\%)  &  73.4  &  56.0  &  17.5  &  62.6 \\ \midrule
Short-term Memory  &  71.5  &  \underline{65.6}  &  29.6  &  \underline{73.6} \\
Long-term Memory  &  69.4  &  60.2  &  \underline{36.1}  &  69.8 \\ \midrule
\rowcolor{gray!20} \textbf{Complete VisMem (Ours)} & \textbf{75.1} & \textbf{69.8} & \textbf{41.4} & \textbf{77.0} \\ \bottomrule
\end{tabular}}
\label{tab:ablation}
 \vspace{-14pt}
\end{table}

%% file: sec/5_conclusion.tex
\section{Conclusion}
To address ``visual processing bottleneck" of VLMs that impairs advanced visual capacities, we propose VisMem in this work, a cognitively inspired framework embedding dynamic latent vision memory, which integrates dual specialized memory formers guided by human patterns, with a non-intrusive memory invocation mechanism. Extensive experiments validate VisMem achieves an obvious performance improvement across various benchmarks, and exhibits strong cross-domain generalization, catastrophic forgetting mitigation, compatibility, and efficient inference, unlocking comprehensive and advanced visual potentials.


%% file: sec/X_suppl.tex
\clearpage
\setcounter{page}{1}
\maketitlesupplementary

\section{Theoretical Foundations}
\label{sec:thoery}
As the mainstream position in anthropological cognitive psychology since the 20th century, short-term memory and long-term memory are two distinct storage systems that can be differentiated based on their functional and neural underpinnings~\cite{baddeley2012working,norris2017short}. Specifically, the \textit{Dennis Norris Theory}~\cite{norris2017short} proposes that short-term memory requires processing new visual information, temporarily storing multiple tokens, and enabling variable signals. It relies neurologically on vision-specific brain regions, \textit{e.g.}, the visual cortex and the posterior superior temporal lobe associated with verbal short-term memory), exhibiting visual dominance; long-term memory, however, centers on abstract semantic representations and relies on semantic-related brain regions like the medial temporal lobe and mid-temporal lobe.

Thus, we propose a framework termed VisMem to invoke dual short and long latent memory during the token-by-token autoregressive generation. Aligned with \textit{Dennis Norris Theory}~\cite{norris2017short}, we instantiate these roles in a VLM backbone via latent vision memory invocation and latent vision memory formation, which together produce distinct short and long latent memory tokens and integrate them into the generation stream of the model.

\section{Methodology Details}
\subsection{Query Builder}
As described in \cref{sec:memory_formation}, the we initialize a lightweight transformer-based encoder as memory builder $\mathcal{B}$. We feed the concatenated memory query $\mathbf{Q}$ and hidden states of vision and output $\mathbf{H}$ into the builder to encoder query as memory hook (see \cref{eq:query_builder}). The transformer-based builder has $L$ layers of encoders, the output process of the $\ell$ layer could be summarized as:
\begin{align}
    &\operatorname{SA}(x)=\operatorname{SM}\left(\frac{\left(x W_{q}\right)\left(x W_{k}\right)^{\top}}{\sqrt{d_{k}}}+M\right)\left(x W_{v}\right), \\
    &x^{\ell}=\operatorname{FF}\left(\operatorname{LN}\left(x^{\ell-1}+\operatorname{SA}\left(\operatorname{LN}\left(x^{\ell-1}\right)\right)\right)\right)+x^{\ell-1}, \label{eq:encoder_layer}
\end{align}
where we simplify the input sequence to $x$, and $\operatorname{SM}$, $\operatorname{MHA}$, $\operatorname{FF}$, $\operatorname{LN}$ denote the softmax, multi-head self-attention, feed-forward layer, layer normalization operations, respectively. In addition, $M$ is the mask which only allows attention from memory query $\mathbf{Q}$ to hidden states $\mathbf{H}$, and blocks the reverse direction:
\begin{equation}
    M_{ij}=
\begin{cases}
-C, & i < K \; and \; j \ge K\\
0, &otherwise 
\end{cases},
\label{eq:mask_attention}
\end{equation}
where $C\gg 0$ is constant, thus the attention is close to $-\infty$.

\subsection{Training Recipe}
\label{sec:training_appendix}
As mentioned in \cref{sec:training}, we design a two-stage training pipeline: at the first stage, the main objective is to optimize the memory formation process (see \cref{eq:stage1_objective}); at the second stage, the main objective is to optimize the memory invocation (see \cref{eq:stage2_objective}). We update the models based on reinforcement learning, \textit{i.e.}, GRPO strategy~\cite{shao2024deepseekmath}. Specifically, for each instruction-vision pair $\left(I,V\right)$, the policy model $\mathcal{P}$ generates a group of $G$ distinct candidate trajectories, termed as $\mathcal{T}=\left\{\tau_{1},\dots,\tau_{G}\right\}$. For each trajectory, we utilize a $S\left(\cdot\right)$ to quantify the performance. Then, a group-relative baseline is calculated via averaging  and standardizing all trajectories within the candidate group $G$:
\begin{equation}
    \overline{S}=\frac{1}{G} \sum_{i=1}^{G}S\left(\tau_{i}\right),\hat{S}=\sqrt{\frac{1}{G} \sum_{i=1}^{G}\left(S\left(\tau_{i}\right)-\overline{S}\right)^{2}}.
\end{equation}

Consequently, the group-relative advantage of each trajectory could be formulated as:
\begin{equation}
    \hat{A} = \frac{S\left(\tau\right)-\overline{S}}{\hat{S}+\epsilon}.
\end{equation}

At the \textbf{Stage I}, the reinforcement learning optimizes the memory formation process, whose final objective function is:
\begin{equation}
\begin{split}
    &\mathcal{J}_{GRPO}^{stage1}(\phi) = \mathbb{E}_{\tau,\mathbf{M}_{s/l},\mathbf{Q}}\left[\frac{1}{G} \sum_{i=1}^{G} \right. \\
    &\quad \min \left(\rho_i(\phi) \hat{A}_{i}, \operatorname{clip}\left(\rho_i(\phi), 1-\epsilon, 1+\epsilon\right) \hat{A}_{i}\right)\Bigg] \\
    &- \beta D_{\mathrm{KL}}\left[\pi^{\phi}_{\tau} \| \pi^{\phi}_{\mathrm{ref}}\right],
\end{split}
\label{eq:stage1_appendix}
\end{equation}
where $\epsilon$ controls the group-relative advantage $\hat{A}$, $\beta$ regulates the KL divergence penalty, and the updated  policy parameters $\pi^{\phi}=\pi^{\phi}\left(\mathbf{Q}\mid\mathbf{H}\right)\cdot\pi^{\phi}\left(\mathbf{M}_{s/l}\mid\mathbf{Q}\right)$.

At the \textbf{Stage II}, the reinforcement learning optimizes the memory invocation process, whose final objective function is:
\begin{equation}
\begin{split}
    &\mathcal{J}_{GRPO}^{stage2}(\theta) = \mathbb{E}_{\tau,x}\left[\frac{1}{G} \sum_{i=1}^{G} \right. \\
    &\quad \min \left(\rho_{i}(\theta) \hat{A}_{i}, \operatorname{clip}\left(\rho_{i}(\theta), 1-\epsilon, 1+\epsilon\right) \hat{A}_{i}\right)\Bigg] \\
    &- \beta D_{\mathrm{KL}}\left[\pi^{\theta}_{\tau} \| \pi^{\theta}_{\mathrm{ref}}\right].
\end{split}
\label{eq:stage2_appendix}
\end{equation}

\section{Experiment Details}
\subsection{Training Data}
During the two-stage training procedure, we use the same training data to optimize both the memory invocation and memory formation in the latent vision memory system. Initially, we include the training split dataset of the selected benchmarks and retain their original data division. For benchmarks without a training phase, we use them solely for evaluation. Additionally, we incorporate the Visual CoT~\cite{shao2024visual} and Mullberry~\cite{yao2024mulberry}, improving the reasoning abilities.

\subsection{Benchmarks}
\label{sec:benchmark_appendix}
To comprehensively evaluate the performance of the selected baselines, we involve 12 benchmarks, consisting of 5 benchmarks of understanding, 4 benchmarks of reasoning, and 3 benchmarks of generation: 
\begin{itemize}
    \item MMStar~\cite{chen2024we} is a high-quality vision-centric benchmark meticulously curated by human experts. This benchmark assesses 6 core capabilities across 18 detailed axes of visual understanding.
    \item MMVet~\cite{yu2024mmvet} establishes 6 core visual understanding capabilities and investigates 16 critical integrations derived from their combinations. It uses an evaluator tailored for open-ended outputs.
    \item MMT~\cite{ying2024mmt} consists of carefully curated multi-choice visual questions, covering 32 core meta-tasks and 162 subtasks within the field of visual understanding.
    \item BLINK~\cite{fu2024blink} reconstructs 14 classic computer vision tasks into multiple-choice questions. Each question is paired with either single or multiple images and supplemented with visual prompting.
    \item MuirBench~\cite{wang2025muirbench} covers 12 diverse multi-image tasks, which involve 10 categories of multi-image relations. Each standard instance is paired with an unanswerable variant that differs only minimally in semantics.
    \item MMMU~\cite{yue2024mmmu} comprises meticulously curated visual questions sourced from college exams, quizzes, and textbooks spanning 30 subjects and 183 subfields, which focus on advanced reasoning grounded in domain-specific knowledge.
    \item LogicVista~\cite{xiao2024logicvista} evaluates general logical cognition abilities across 5 logical reasoning tasks, which encompass 9 distinct capabilities. Each question is annotated with the correct answer and the human-written reasoning behind the selection.
    \item MathVista~\cite{wang2024measuring} unifies the challenges of heterogeneous mathematical and visual tasks, which are curated from math-oriented multimodal datasets.
    \item MV-Math~\cite{wang2025mv} is a dataset comprising mathematical problems, integrating multiple images interleaved with text, and detailed annotations. It features multiple-choice, free-form, and multi-step questions across 11 subject areas at 3 difficulty levels.
    \item HallBench~\cite{guan2024hallusionbench} consists of images paired with questions, designed by human experts to assess the hallucination level of generation.
    \item MultiTrust~\cite{zhang2024multitrust} covers five primary aspects: truthfulness, safety, robustness, fairness, and privacy,  evaluating the trustworthiness of generation.
    \item MMVU~\cite{liu2025unveiling} encompasses 12 categories, and designs evaluation metrics that measure the quality and error degree of generation.
\end{itemize}

\begin{table*}[t]
\centering
\caption{Configurations of parameters.}
\setlength{\tabcolsep}{0.9mm}
\resizebox{0.55\textwidth}{!}{
\begin{tabular}{l|ccc}
\toprule   
\textbf{Configurations} & \textbf{Parameters} & \multicolumn{2}{c}{\textbf{Values}} \\ \midrule
\multirow{3}{*}{Core} & $K$ & \multicolumn{2}{c}{8} \\
& $N_s$ & \multicolumn{2}{c}{4}\\
& $N_l$ & \multicolumn{2}{c}{8}\\ \midrule
\multirow{4}{*}{LoRA~\cite{hu2022lora}} & $rank$ & \multicolumn{2}{c}{16} \\
& $\alpha$ & \multicolumn{2}{c}{32} \\
& $drop\_out\_rate$ & \multicolumn{2}{c}{0.1} \\
& $target\_module$ & \multicolumn{2}{c}{$[q\_proj,v\_proj]$} \\ \midrule
\multirow{12}{*}{Training} & & \textbf{Stage I} & \textbf{Stage II} \\
& $batch\_size$ & \multicolumn{2}{c}{8} \\
& $epoch$ & \multicolumn{2}{c}{2} \\
& $warmup\_ratio$ & 0.2 & 0.1 \\
& $num\_iteration$ & \multicolumn{2}{c}{1} \\
& $learning\_rate$ & $5e^{-5}$ & $1e^{-5}$\\
& $optimizer$ & \multicolumn{2}{c}{AdamW~\cite{loshchilov2017decoupled}} \\
& $scheduler$ & \multicolumn{2}{c}{Cosine} \\
& $group\_size$ & \multicolumn{2}{c}{16} \\
& $clip\_ratio$ & \multicolumn{2}{c}{0.2} \\
& $kl\_penalty\_coefficient\;\beta$ & 0.015 & 0.030 \\
& $target\_kl\_per\_token$ & 0.03 & 0.05 \\
& $penalty\_intensity\;\alpha$ & - & 0.3 \\
\bottomrule
\end{tabular}}
\label{tab:implementation_appendix}
\end{table*}

\begin{table*}[t]
\centering
\caption{Results on 9 selected subsets of MuirBench~\cite{wang2025muirbench}. We compare our VisMem with the second and third best scored counterparts, and separately use the short or long latent memory to assess the improvements of each.}
\setlength{\tabcolsep}{0.9mm}
\resizebox{1\textwidth}{!}{
\begin{tabular}{l|ccccccccc}
\toprule   
\multirow{1}{*}{\textbf{Method}} & \textbf{Counting} & \textbf{Grounding} & \textbf{Matching} & \textbf{Scene} & \textbf{Difference} & \textbf{Cartoon} & \textbf{Diagram} & \textbf{Geographic} & \textbf{Retrieval}  \\ \midrule
Vanilla~\cite{bai2025qwen2} & 44.1 & 34.2 & 80.9 & 70.5 & 53.2 & 52.9 & 82.4 & 53.7 & 76.1 \\ \midrule
VLM-R1~\cite{shen2025vlm} & 52.5 & 38.1 & 83.6 & 73.5 & 58.1 & 55.1 & 86.8 & 56.7 & 79.4 \\
Vision-R1~\cite{huang2025vision} & 53.8 & 39.2 & \textbf{84.5} & 73.1 & 57.4 & 57.2 & 87.4 & 57.9 & 78.9 \\ \midrule
VisMem (Short Memory) & \textbf{61.3} & \underline{49.4} & 82.7 & 72.1 & \underline{58.9} & 54.0 & \underline{88.9} & 61.8 & \underline{87.5} \\
VisMem (Long Memory) & 46.3 & 42.6 & 83.2 & \underline{74.3} & 55.4 & \underline{59.4} & 87.4 & \underline{62.7} & 78.3 \\ \midrule
VisMem & \underline{60.8} & \textbf{52.3} & \underline{84.0} & \textbf{76.2} & \textbf{60.6} & \textbf{59.7} & \textbf{90.1} & \textbf{65.5} & \textbf{89.8} \\ \bottomrule
\end{tabular}}
\label{tab:muirbench}
\end{table*}

\begin{table*}[t]
\centering
\caption{Results on 10 selected subsets (5 reasoning skills and 5 capabilities) of LogicVista~\cite{xiao2024logicvista}. We compare our VisMem with the second and third best scored counterparts, and separately use the short or long latent memory to assess the improvements of each.}
\setlength{\tabcolsep}{0.9mm}
\resizebox{1\textwidth}{!}{
\begin{tabular}{l|ccccc|ccccc}
\toprule         
\multirow{1}{*}{\textbf{Method}} & \textbf{Inductive} & \textbf{Deductive} & \textbf{Numerical} & \textbf{Spatial} & \textbf{Mechanical}  & \textbf{Patterns} & \textbf{Puzzles} & \textbf{OCR} & \textbf{Graphs} & \textbf{Tables} \\ \midrule
Vanilla~\cite{bai2025qwen2} & 44.6 & 45.0 & 39.7 & 37.9 & 48.7 & 30.1 & 32.5 & 41.6 & 34.4 & 36.8\\ \midrule
VLM-R1~\cite{shen2025vlm} & 53.7 & 52.7 & 45.8 & 44.1 & 57.3 & 35.8 & 42.8 & \underline{49.0} & 46.5 & 52.6 \\
Vision-R1~\cite{huang2025vision} & 53.5 & 51.4 & \underline{46.7} & 44.8 & \textbf{58.9} & \underline{36.5} & \underline{43.6} & \textbf{49.7} & 48.2 & 53.8 \\ \midrule
VisMem (Short Memory) & 49.8 & 50.1 & 44.7 & \underline{45.2} & 54.3 & 35.2 & 42.0 & 47.6 & \underline{50.3} & \underline{54.1}  \\
VisMem (Long Memory)  & \underline{57.5} & \underline{58.4} & 42.8 & 40.0 & 52.0 & 35.7 & 38.0 & 47.4 & 48.9 & 51.3 \\ \midrule
VisMem & \textbf{59.4} & \textbf{59.8} & \textbf{46.9} & \textbf{47.2} & \underline{57.4} & \textbf{38.9} & \textbf{44.6} & 48.5 & \textbf{52.8} & \textbf{57.9} \\ \bottomrule
\end{tabular}}
\label{tab:logic_vista}
\end{table*}

\subsection{Baselines}
\label{sec:baseline_appendix}

We select a total of 16 baselines, including the vanilla model~\cite{bai2025qwen2}, 5 direct training paradigms: SFT, Visual-RFT~\cite{liu2025visual}, VLM-R1~\cite{shen2025vlm}, Vision-R1~\cite{huang2025vision}, and PAPO~\cite{wang2025perception}; 5 image-level paradigms: Sketchpad~\cite{hu2024visual}, GRIT~\cite{fan2025grit}, PixelReasoner~\cite{su2025pixel}, DeepEyes~\cite{zheng2025deepeyes}, and OpenThinkImg~\cite{su2025openthinkimg}; 4 token-level paradigms: Scaffold~\cite{lei2024scaffolding}, ICoT~\cite{gao2025interleaved}, MINT-CoT~\cite{chen2025mint}, and VPT~\cite{yu2025introducing}; and 1 latent space paradigm: Mirage~\cite{yang2025machine}.

Here, VLM-R1~\cite{shen2025vlm} and Vision-R1~\cite{huang2025vision} follow the main GRPO~\cite{guo2025deepseek} paradigm based on VLMs. To assess the effectiveness of different methods, our VisMem is trained on Qwen-2.5-VL-7B~\cite{bai2025qwen2}. For strategies initially implemented on other base models, \textit{e.g.}, GPT-4o~\cite{hurst2024gpt} and Qwen2-VL~\cite{wang2024qwen2}, we transfer them to Qwen2.5-VL-7B~\cite{bai2025qwen2} for fair comparison. Besides, we maintain identical training datasets across most counterparts; however, for those three methods with specially curated datasets, we follow their original settings. Namely, Mirage~\cite{yang2025machine} requires additional labeled training images, so we follow its original training dataset; GRIT~\cite{fan2025grit} uses a tailored training process with designed data; and MINT-CoT~\cite{chen2025mint} curates high-quality mathematical samples with grids and annotations.

\subsection{Implementations}
\label{sec:implementation_appendix}
The configurations and implementations of the experiments include three main parts: the core hyperparameters, the parameters of the LoRA adapter, and the parameters we use during training. The configurations and implementations of the experiments are listed in \cref{tab:implementation_appendix}.

\section{Additional Results}
\subsection{Benchmark Subset Results towards Visual Sub-capacities}
To precisely identify the capability boundaries and advantages of our VisMem, rather than relying solely on overall scores to judge its quality, we evaluate the results of subsets of MuirBench~\cite{wang2025muirbench} and LogicVista~\cite{xiao2024logicvista} benchmarks. We select 9 subsets of the former benchmark, including: counting, grounding, matching, scene, difference, cartoon, diagram, geographic, and retrieval. While in the latter benchmark, we also select 10 subsets, including 5 reasoning skills: inductive, deductive, numerical, spatial, and mechanical, and 5 capacities: patterns, puzzles, OCR, graphs, and tables. It is worth noting that the selected subsets are only part of the benchmark, thus, the average values of the 10 subsets are not the results of the benchmarks.

As listed in \cref{tab:muirbench}, compared with VLM-R1~\cite{shen2025vlm} and Vision-R1~\cite{huang2025vision}, our VisMem achieves the best results on 7 subsets and ranks second on the remaining two subsets. Specifically, it has a generalized enhancement of at least 5\% over the base model. Besides, VisMem improves the performance the vanilla model by 16.7\% / 18.2\% / 11.8\% / 13.7\% on the counting, grounding, geographic, and retrieval sub-tasks, vastly exceeding the second-best counterpart by 7.0-13.1\%. These results indicate that our latent vision memory system significantly promote the fine-grained visual cognition and perception of the base VLMs.

As presented in \cref{tab:logic_vista}, our VisMem outperforms two baseline models, \textit{i.e.}, VLM-R1~\cite{shen2025vlm} and Vision-R1~\cite{huang2025vision}, by achieving the top performance across 8 subsets. Specifically, it delivers a generalized improvement of no less than 7\% over the base model. Notably, on inductive, deductive, graph-based, and table-based sub-tasks, VisMem surpasses the vanilla model by 14.8\%, 14.8\%, 18.4\%, and 21.1\%, respectively, which exceeds the second-ranked model by a substantial margin of 5.3–7.1\%. These results demonstrate that our latent visual memory system delivers contextualized semantic knowledge, thereby enhancing visual reasoning and robust generation capabilities.

\begin{table*}[t]
\centering
\caption{Results of various models with full training datasets and partial datasets (Visual CoT~\cite{shao2024visual} and Mulberry~\cite{yao2024mulberry}), and evaluated across four benchmarks.}
\setlength{\tabcolsep}{0.9mm}
\resizebox{0.57\textwidth}{!}{
\begin{tabular}{l|cccccccc}
\toprule   
\multirow{2}{*}{\textbf{Method}} & \multicolumn{2}{c}{\textbf{MMVet}} & \multicolumn{2}{c}{\textbf{MuirBench}} & \multicolumn{2}{c}{\textbf{MV-Math}} & \multicolumn{2}{c}{\textbf{MultiTrust}}  \\ \cmidrule(lr){2-3} \cmidrule(lr){4-5} \cmidrule(lr){6-7} \cmidrule(lr){8-9}
 & Full & Part &  Full & Part & Full & Part & Full & Part  \\ \midrule
Vanilla~\cite{bai2025qwen2}  &  \multicolumn{2}{c}{66.0} &  \multicolumn{2}{c}{57.4} &  \multicolumn{2}{c}{18.9} &  \multicolumn{2}{c}{64.8} \\ \midrule
SFT & 67.5 & 65.8 & 58.7 & 57.2 & 22.8 & 21.2 & 67.0 & 65.4 \\
Visual-RFT~\cite{liu2025visual} & 70.5 & 65.3 & 62.9 & 57.8 & 26.5 & 24.2 & 70.7 & 66.0 \\
VLM-R1~\cite{shen2025vlm} & \underline{73.0} & 67.7 & 63.8 & 59.0 & 34.6  & 32.1 & 69.9 & 66.1 \\
Vision-R1~\cite{huang2025vision} & 71.7 & 68.4 & \underline{64.0} & 59.8 & 38.7 & 35.6 & 72.6 & 67.1 \\ 
PAPO~\cite{wang2025perception} & 69.8 & 68.6 & 56.7 & 56.4 & 34.8 & 32.8 & 67.7 & 66.4 \\\midrule
DeepEyes~\cite{zheng2025deepeyes} & 70.5 & 67.9  &  63.0 & \underline{60.6} & 31.5  & 27.9 & 72.6 & 68.5 \\
OpenThinkImg~\cite{su2025openthinkimg} & 71.6 & \underline{69.5} & 61.7 & 59.7 &  28.0 & 25.9 & \underline{74.0} & 68.3 \\\midrule
ICoT~\cite{gao2025interleaved} & 67.9 & 67.1 & 57.0 & 56.4 & 30.8 & 28.3 & 69.1 & 68.4 \\
MINT-CoT~\cite{chen2025mint} & 69.5 & 68.4 & 58.9 & 57.8 &  \underline{39.2} & \underline{36.4} &  71.4 & \underline{70.2} \\ \midrule
Mirage~\cite{yang2025machine} & 71.8 & 70.2 & 59.0 & 57.2 & 35.4 & 33.1 &  66.1 & 64.0 \\
\rowcolor{gray!20} \textbf{VisMem (Ours)} & \textbf{75.1} & \textbf{72.9} &  \textbf{69.8} & \textbf{66.4} & \textbf{41.4} & \textbf{39.1} & \textbf{77.0} & \textbf{74.9} \\ \bottomrule
\end{tabular}}
\label{tab:cross_domain}
\end{table*}

\begin{table*}[t]
\centering
\caption{Results of various models on MMVet~\cite{yu2024mmvet} with four-stage continual learning. Stage 0: MMVet~\cite{yu2024mmvet}; Stage 1: BLINK~\cite{fu2024blink}, and MuirBench~\cite{wang2025muirbench}; Stage 2: LogicVista~\cite{xiao2024logicvista}, and Math-V~\cite{wang2024measuring}; Stage 3: MultiTrust~\cite{zhang2024multitrust}, and MMVU~\cite{liu2025unveiling}.}
\setlength{\tabcolsep}{0.9mm}
\resizebox{0.57\textwidth}{!}{
\begin{tabular}{l|cccc|c}
\toprule   
\multirow{1}{*}{\textbf{Method}} & \textbf{Stage 0} & \textbf{Stage 1} & \textbf{Stage 2} & \textbf{Stage 3} & \textbf{Original} \\ \midrule
Vanilla~\cite{bai2025qwen2}  &  \multicolumn{5}{c}{66.0} \\ \midrule
SFT & 71.4 & 70.6 & 62.3 & 60.1 & 67.5  \\
Visual-RFT~\cite{liu2025visual} & 74.0 & 72.2 & 67.3 & 65.7 & 70.5 \\
VLM-R1~\cite{shen2025vlm} & 77.8 & 74.1 & 66.4 & 66.9 & \underline{73.0} \\
Vision-R1~\cite{huang2025vision} & 76.9 & 74.0 & 66.1 & 66.3 & 71.7 \\ 
PAPO~\cite{wang2025perception} & 75.0 & 74.5 & 63.4 & 62.9 & 69.8  \\\midrule
DeepEyes~\cite{zheng2025deepeyes} & 74.1 & 74.6 & \underline{68.9} & \underline{68.4} & 70.5 \\
OpenThinkImg~\cite{su2025openthinkimg} & 76.2 & 74.7 & 66.5 &  67.9 & 71.6 \\\midrule
ICoT~\cite{gao2025interleaved} & 71.9 & 71.3 & 67.1 & 64.7 & 67.9 \\
MINT-CoT~\cite{chen2025mint} & 72.4 & 71.8 & 65.8 & 66.2 & 69.5 \\ \midrule
Mirage~\cite{yang2025machine} & \textbf{79.1} & \underline{77.8} & 68.7 & 67.0 & 71.8 \\
\rowcolor{gray!20} \textbf{VisMem (Ours)} & \underline{78.6} & \textbf{78.9} & \textbf{71.3} & \textbf{72.1} & \textbf{75.1} \\ \bottomrule
\end{tabular}}
\label{tab:catastrophic_forgetting}
\end{table*}

\begin{figure*}[t]
    \centering
    \includegraphics[width=0.75\linewidth]{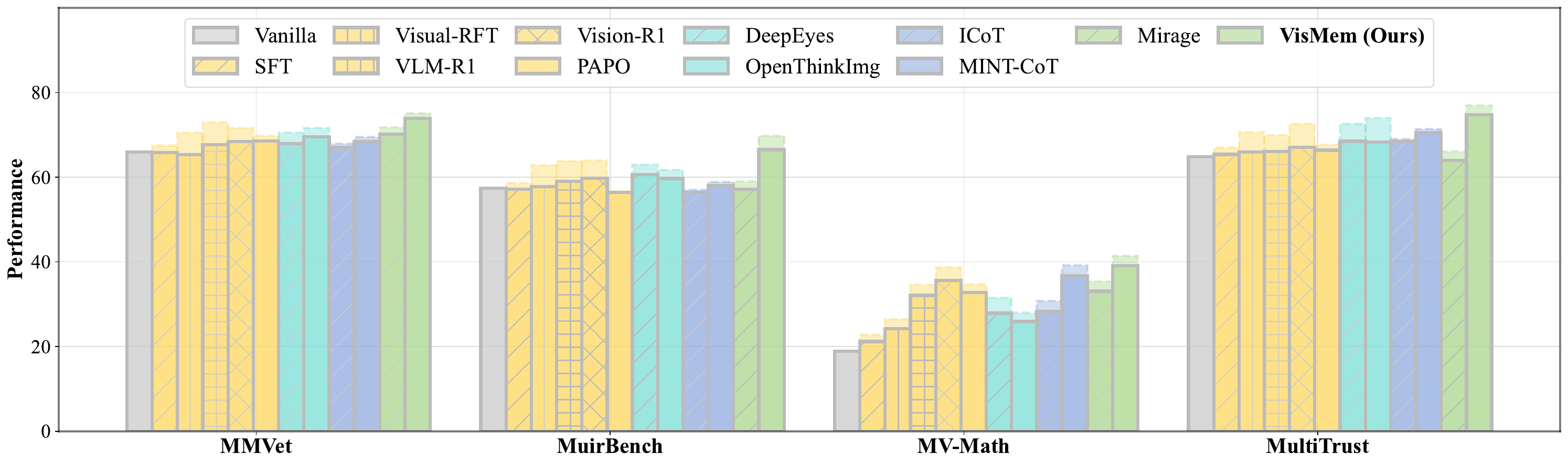}
    \caption{Results of various models of the cross-domain generalization study. Models are only trained on Visual CoT~\cite{shao2024visual} and Mulberry~\cite{yao2024mulberry}, and are evaluated on four benchmarks.}
    \label{fig:cross_domain_appendix}

    \vspace{10pt}
    \includegraphics[width=0.75\linewidth]{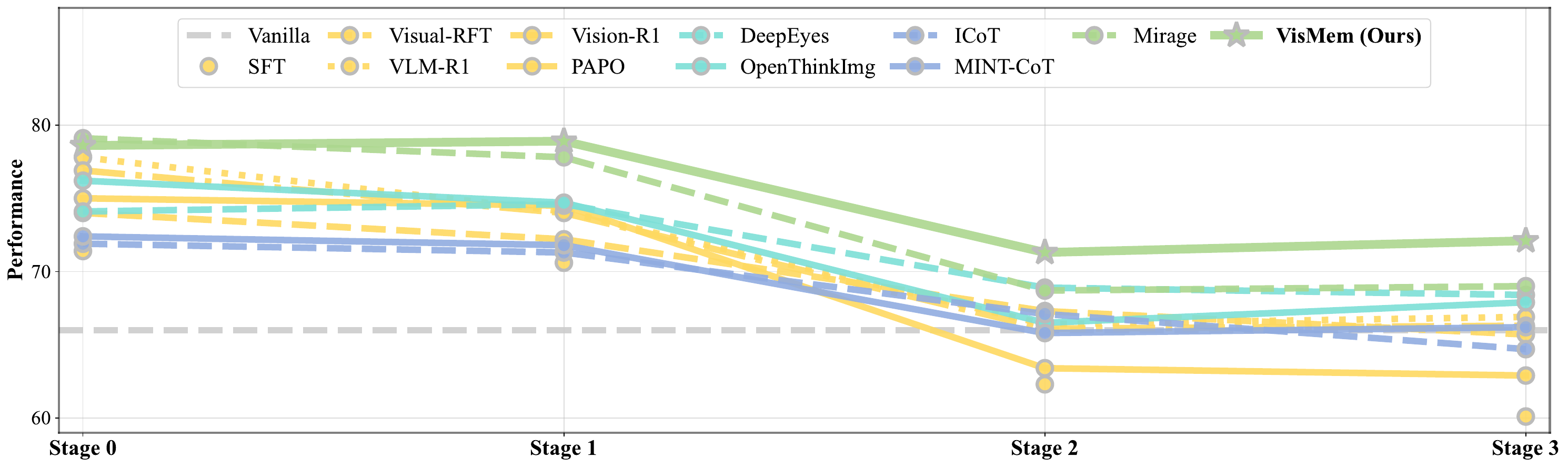}
    \caption{Results of four-stage continual learning on MMVet~\cite{yu2024mmvet}. The model is sequentially trained on
each training data combination (Stage 0 $\to$ Stage 1 $\to$ Stage 2 $\to$ Stage 3). Stage 0 only includes MMVet~\cite{yu2024mmvet} as training data, while Stage 1, 2, 3 add data targeting visual understanding~\cite{fu2024blink,wang2025muirbench}, reasoning~\cite{xiao2024logicvista,wang2024measuring}, and generation~\cite{zhang2024multitrust,liu2025unveiling}.}
    \label{fig:forget_appendix}
\end{figure*}

\subsection{Cross-domain Generalization}
To evaluate the cross-domain generalization capability of our model, we train it exclusively on general datasets, namely, Visual CoT~\cite{shao2024visual} and Mullberry~\cite{yao2024mulberry}), to verify whether latent visual memory can be transferred to unseen domains. As shown in \cref{tab:cross_domain} and \cref{fig:cross_domain_appendix}, our method demonstrates superior performance, which exhibits a smaller performance drop than the fully trained model across all four selected benchmarks, confirming strong cross-domain generalization. Despite being trained on only two datasets, our method achieves a significant performance improvement of 9.1–20.5\% across the four benchmarks, with a mere 2\% performance gap relative to the fully trained model. When compared to other baselines, it still maintains a performance lead of 3.4\% / 6.7\% / 2.7\% /  4.7\% across the four evaluations, respectively.

In general, the image-level, token-level, and latent space paradigms suffer from smaller performance degradation, whereas the direct training paradigm exhibits inferior generalization ability. For example, VLM-R1~\cite{shen2025vlm} experiences a 5.3\% performance drop; by contrast, this value is only 2.1\% for OpenThinkImg~\cite{su2025openthinkimg}, 1.1\% for MINT-CoT~\cite{chen2025mint}, and 2.3\% for our method. These results indicate that while direct training optimizations notably improve performance on specific tasks, they compromise generalization ability to some extent.

\subsection{Catastrophic Forgetting Mitigation}
To assess the extent of catastrophic forgetting, we conducted continual learning experiments with our VisMem and other baselines. As presented in \cref{tab:catastrophic_forgetting} and \cref{fig:forget_appendix}, our method effectively mitigates forgetting of earlier tasks. It consistently achieves the best performance at each stage, demonstrating strong robustness against catastrophic forgetting. Following four-stage sequential continual training, it retains 72.1\% performance on MMVet~\cite{yu2024mmvet}, outperforming 68.4\% of DeepEyes~\cite{zheng2025deepeyes} and 67.0\% of Mirage~\cite{yang2025machine}.

While the direct training paradigm significantly improves performance on specific tasks, it adapts to new tasks via direct updates to core parameters. This introduces conflicts when parameter update directions contradict the storage of existing knowledge, compounded by a lack of constraints from prior knowledge. Consequently, in stage 3, the performance of most direct training methods even falls below that of the vanilla model. In contrast, methods such as OpenThinkImg~\cite{su2025openthinkimg} and our proposed VisMem exhibit stronger knowledge retention and forward transfer capabilities. For instance, in stage 3, training on additional datasets further improves their performance on MMVet~\cite{yu2024mmvet}.

\subsection{Versatility across Various Base Models}
\begin{figure*}[t]
    \centering 
    \begin{subfigure}[b]{0.22\textwidth}
        \centering
        \includegraphics[width=\textwidth]{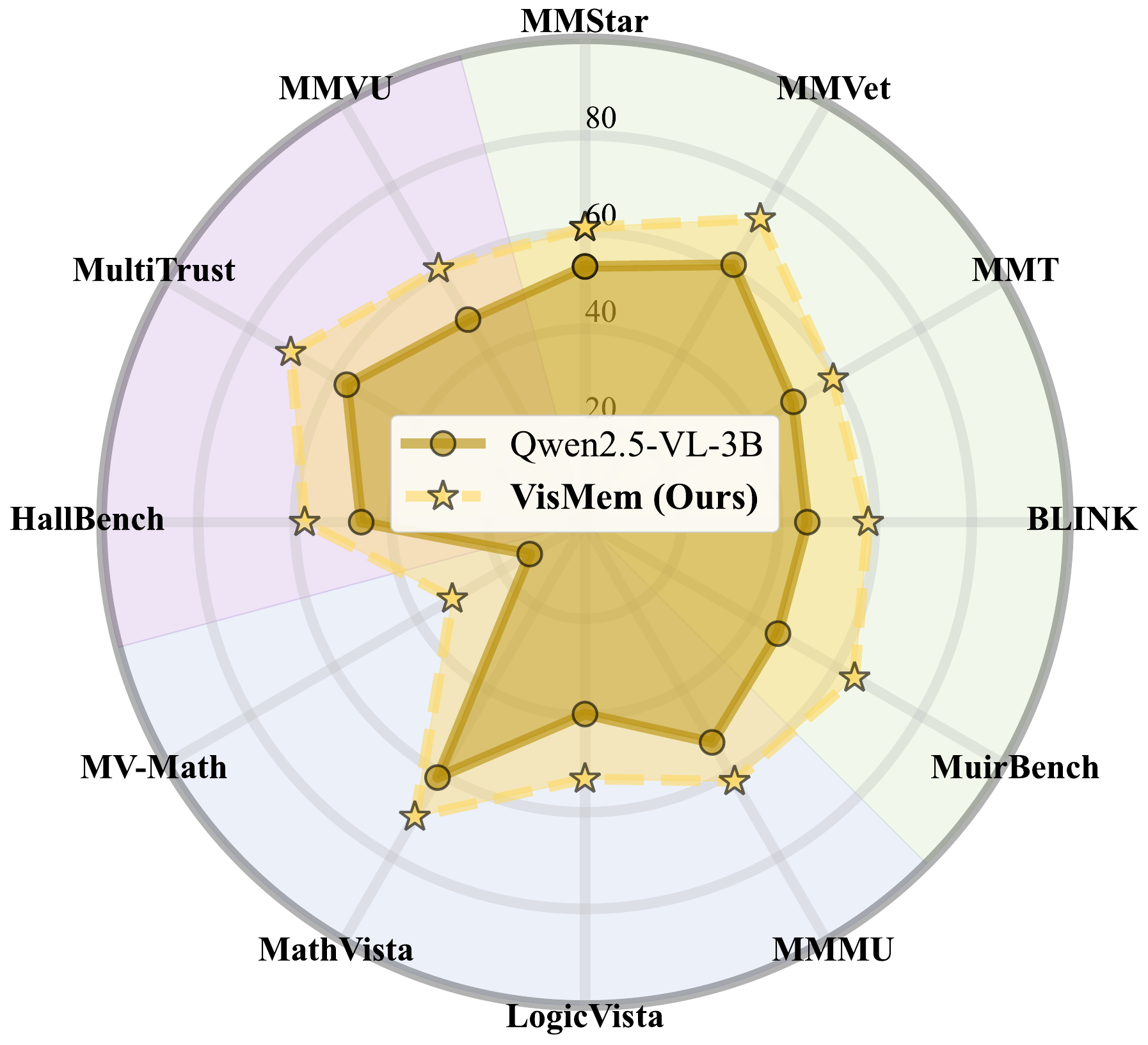} 
        \caption{Qwen2.5-VL-3B} 
    \end{subfigure}  
    \begin{subfigure}[b]{0.22\textwidth}
        \centering
        \includegraphics[width=\textwidth]{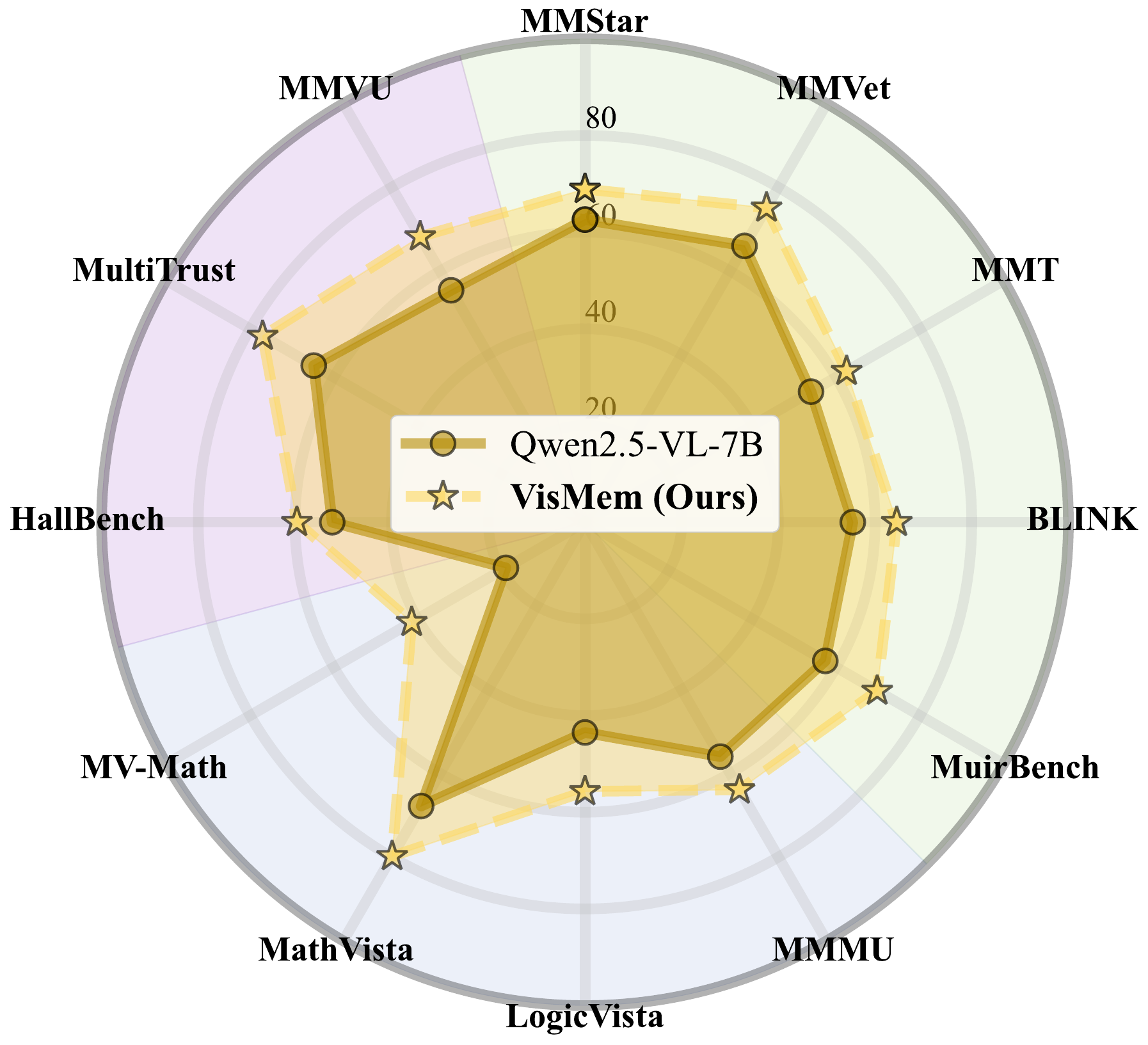}
        \caption{Qwen2.5-VL-7B}  
    \end{subfigure}
    \begin{subfigure}[b]{0.22\textwidth}
        \centering
        \includegraphics[width=\textwidth]{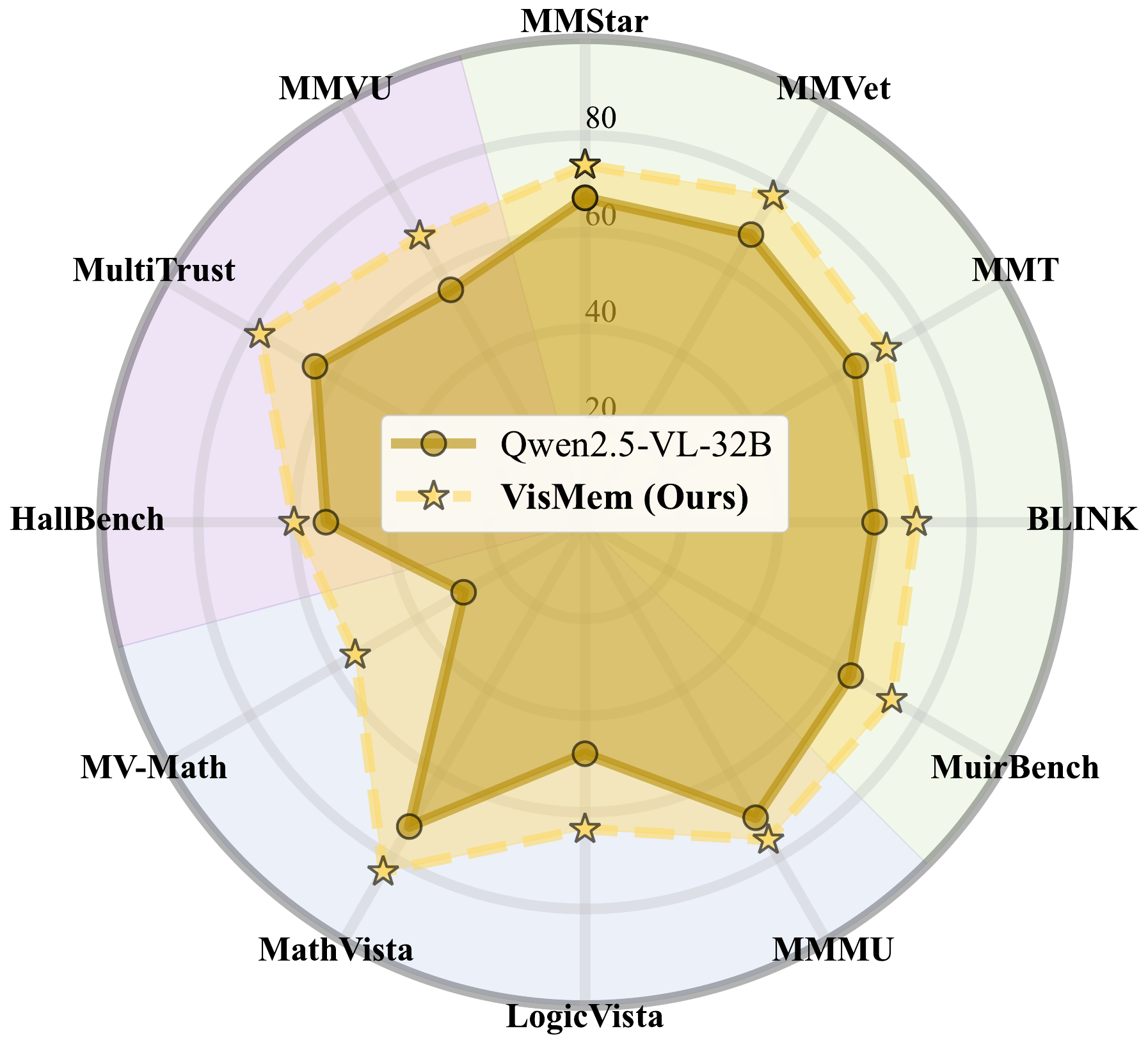}
        \caption{Qwen2.5-VL-32B}  
    \end{subfigure}

    \begin{subfigure}[b]{0.22\textwidth}
        \centering
        \includegraphics[width=\textwidth]{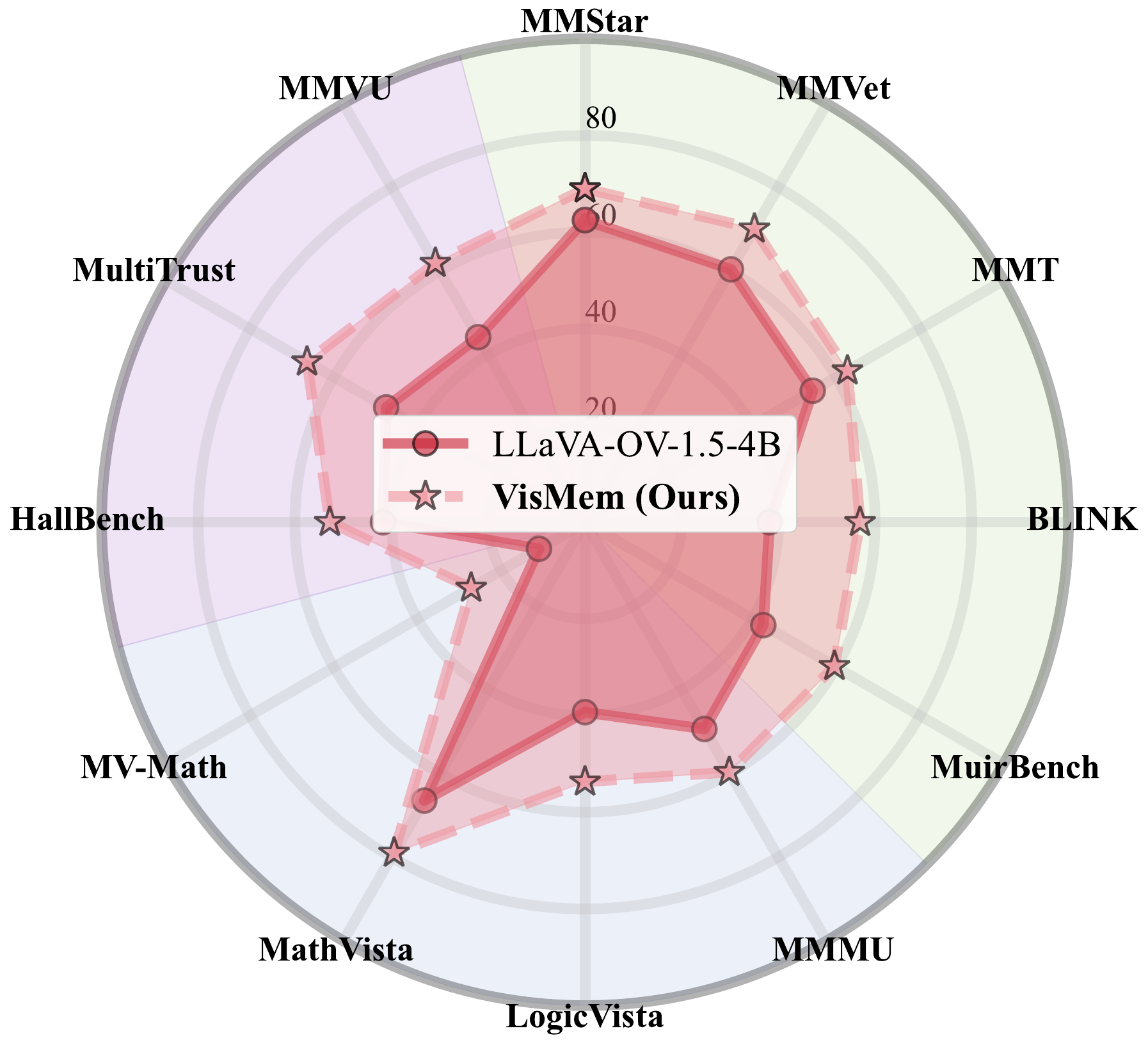}
        \caption{LLaVA-OV-1.5-4B} 
    \end{subfigure}
    \begin{subfigure}[b]{0.22\textwidth}
        \centering
        \includegraphics[width=\textwidth]{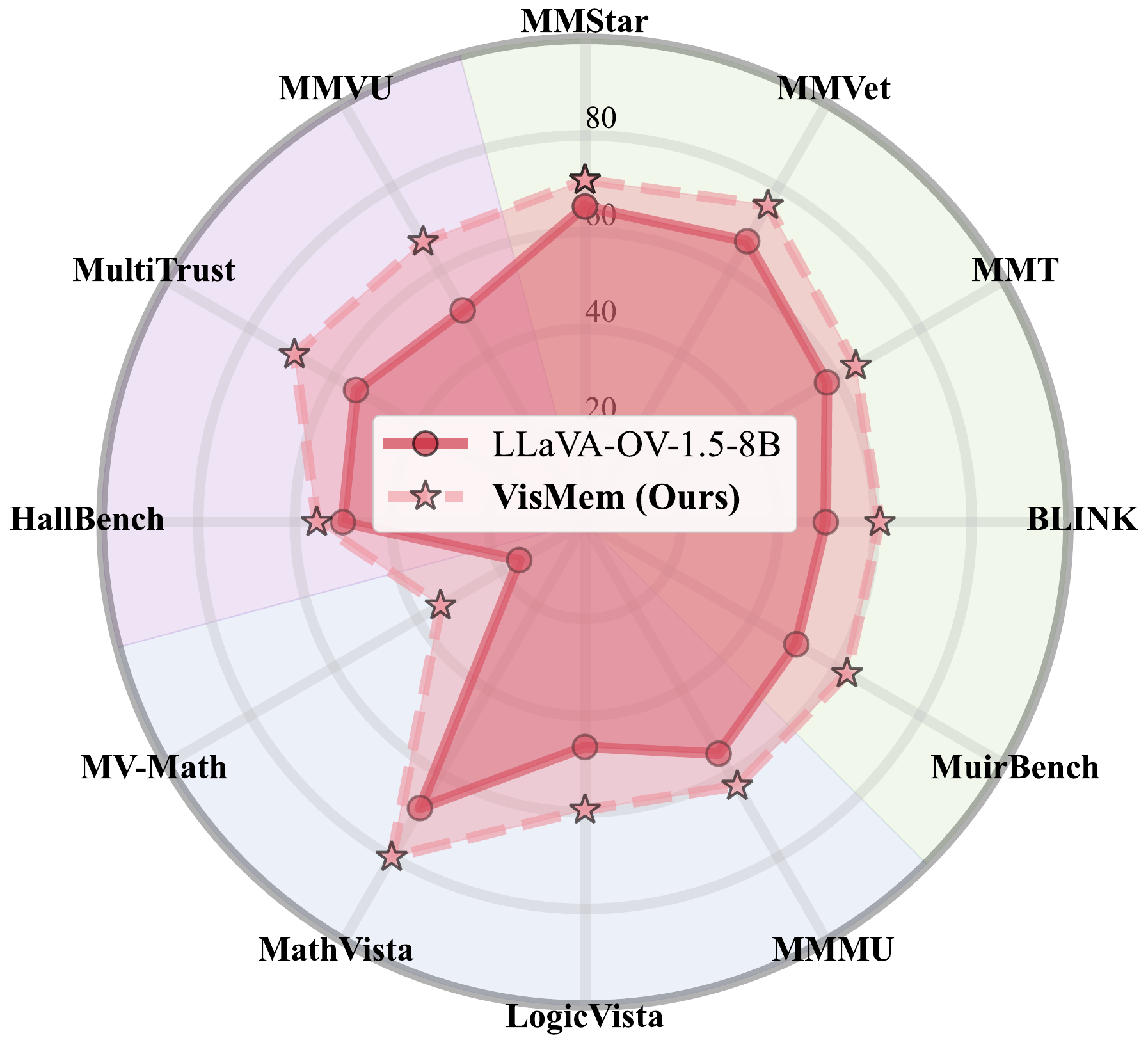}
        \caption{LLaVA-OV-1.5-8B} 
    \end{subfigure}
    \begin{subfigure}[b]{0.22\textwidth}
        \centering
        \includegraphics[width=\textwidth]{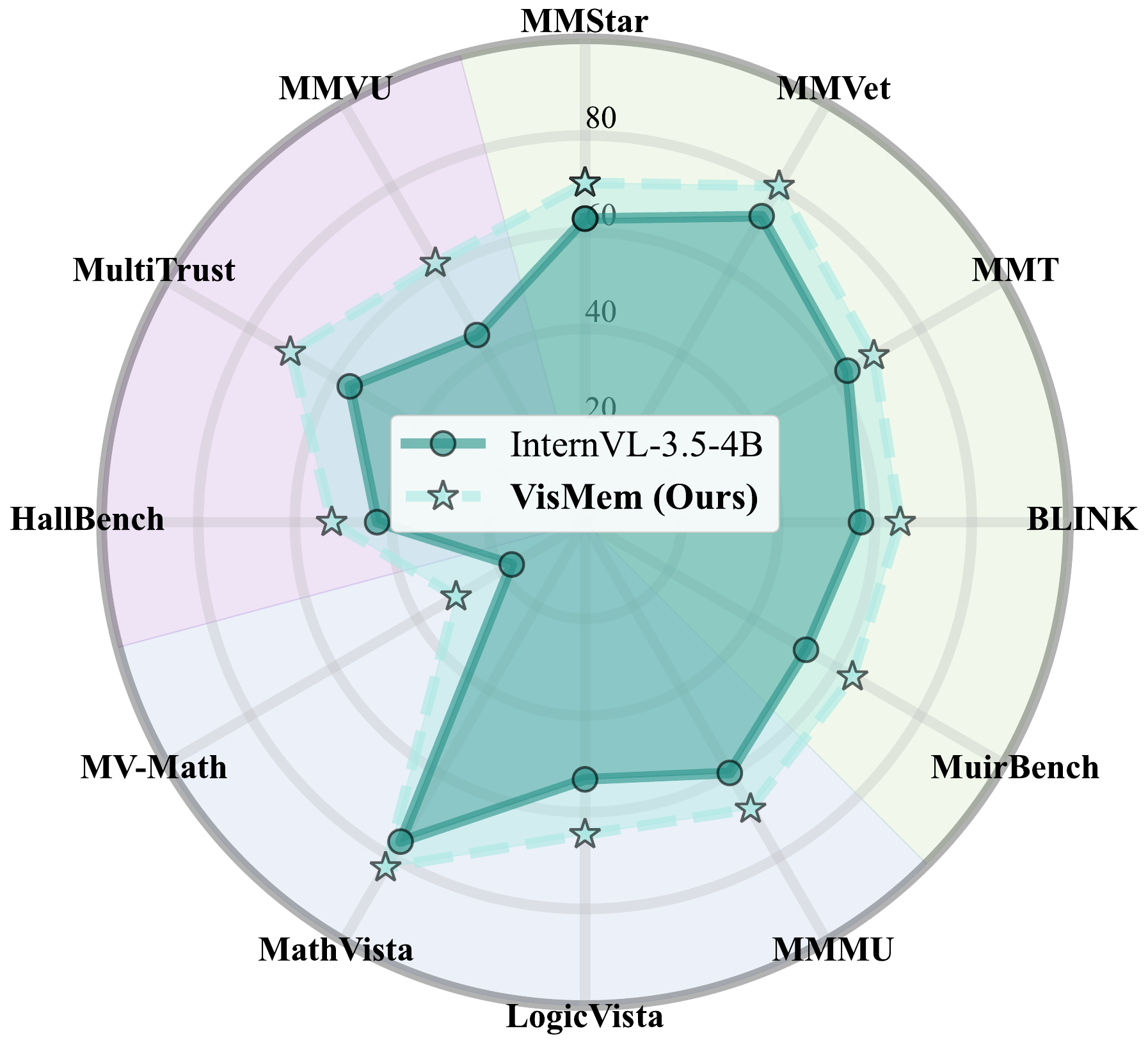}
        \caption{InternVL-3.5-4B}  
    \end{subfigure}

    \begin{subfigure}[b]{0.22\textwidth}
        \centering
        \includegraphics[width=\textwidth]{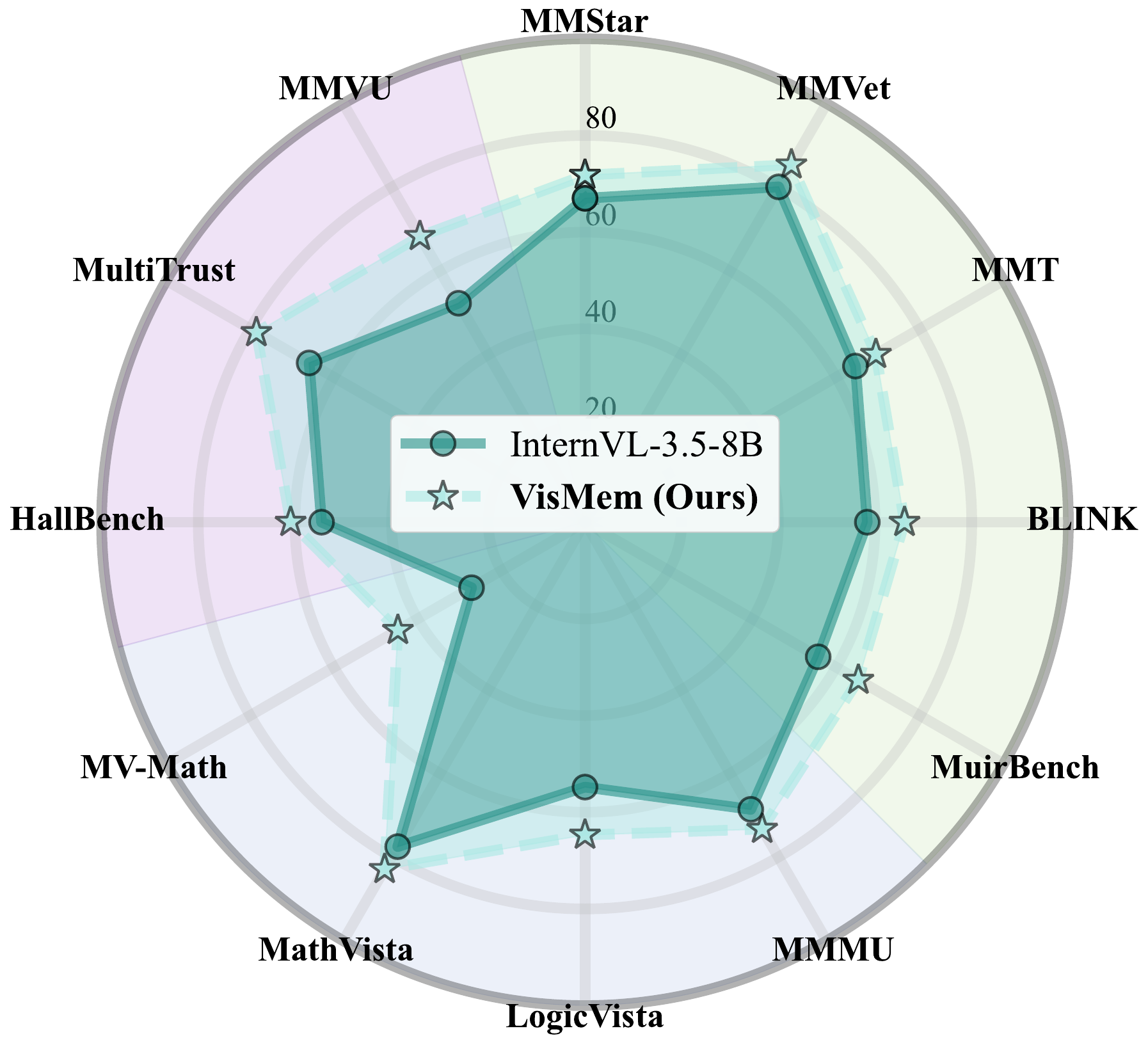}
        \caption{InternVL-3.5-8B} 
    \end{subfigure}
    \begin{subfigure}[b]{0.22\textwidth}
        \centering
        \includegraphics[width=\textwidth]{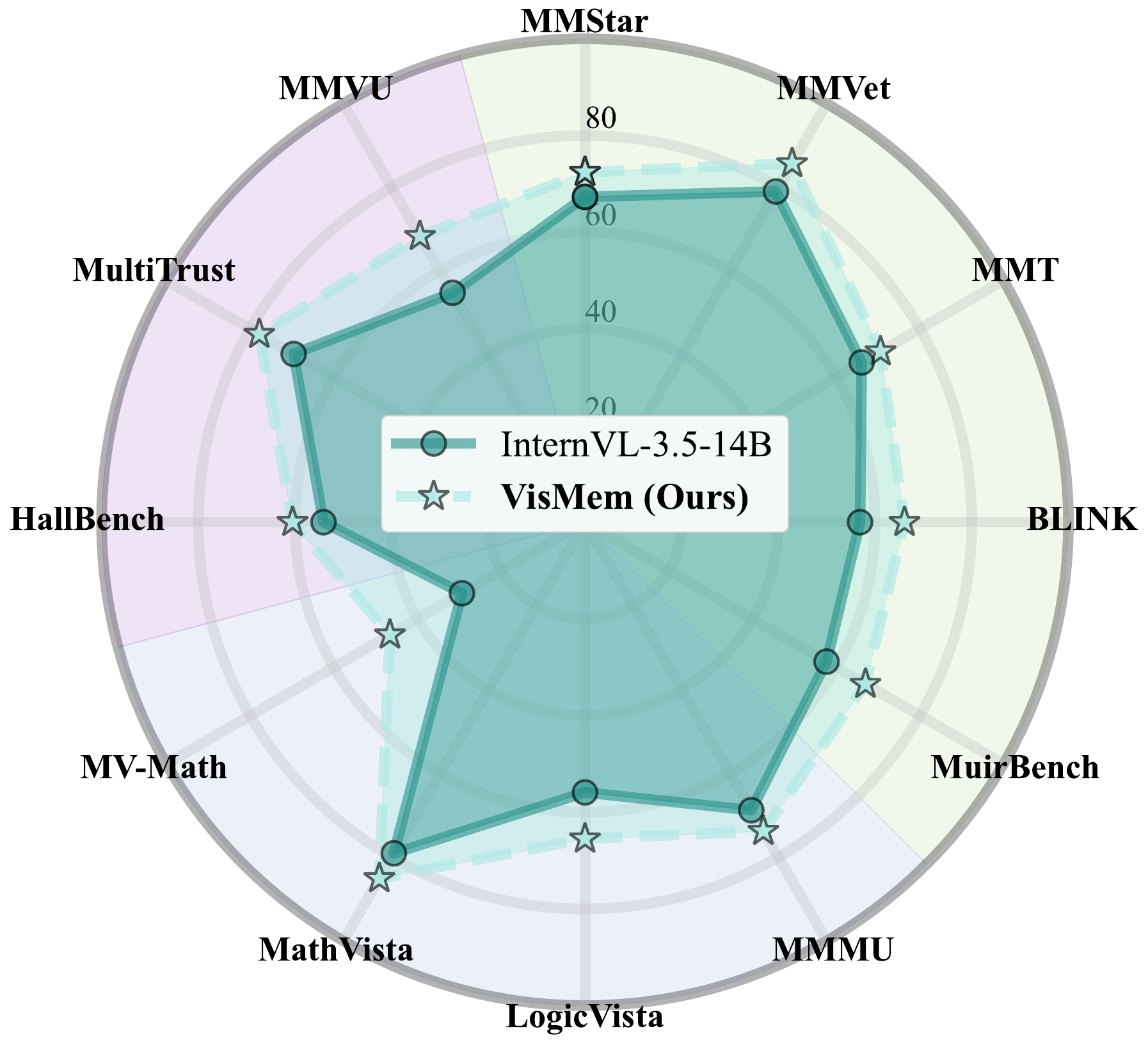}
        \caption{InternVL-3.5-14B} 
    \end{subfigure}
    \begin{subfigure}[b]{0.22\textwidth}
        \centering
        \includegraphics[width=\textwidth]{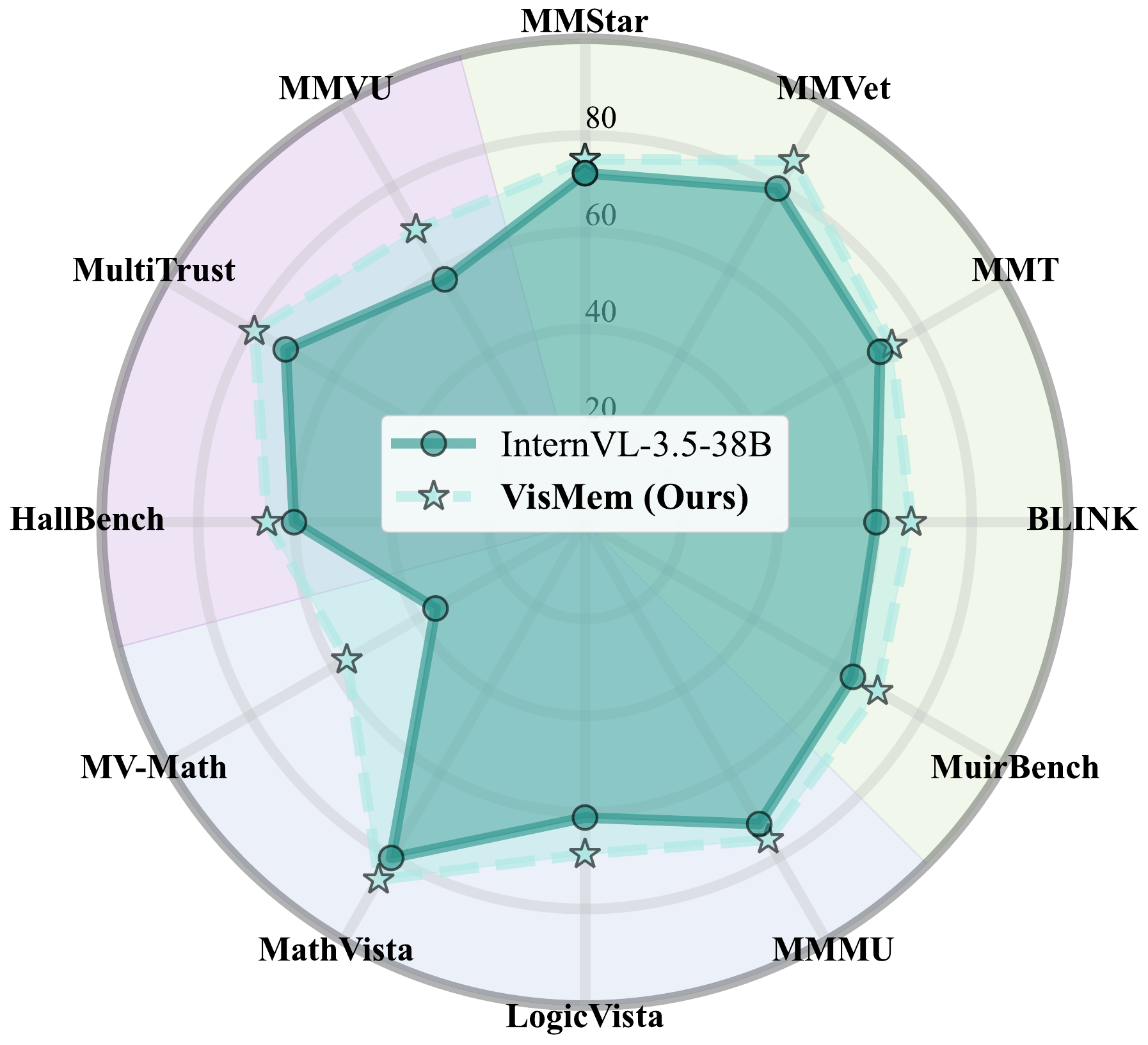}
        \caption{InternVL-3.5-38B}  
    \end{subfigure}
    \caption{Results on different base models.} 
    \label{fig:base_model_appendix} 
\end{figure*}

As presented in \cref{tab:base_models} and \cref{fig:base_model_appendix}, we incorporate our latent visual memory paradigm into 9 base models, including Qwen2.5-VL-3B/7B/32B~\cite{bai2025qwen2}, LLaVA-OV-1.5-4B/8B~\cite{an2025llava}, and InternVL-3.5-4B/8B/14B/38B~\cite{wang2025internvl3}. Our VisMem consistently enhances the visual capabilities of all base models, spanning 3B to 38B parameter sizes across three VLM families. For the widely used medium-sized models (\textit{i.e.}, 7B or 8B parameter models), our latent visual memory delivers substantial performance gains, which brings a 6.3–23.1\% improvement across all benchmarks for Qwen2.5-VL-7B~\cite{bai2025qwen2}, a 5.5–20.2\% improvement for LLaVA-OV-1.5-8B~\cite{an2025llava}, and a 4.8–17.6\% improvement for InternVL-3.5-8B~\cite{wang2025internvl3}, respectively.

Furthermore, in most benchmarks, smaller-parameter base models yield greater performance gains than their medium- or large-sized counterparts. This phenomenon may stem from an imbalance in task difficulty, which makes it more challenging for models with higher baseline scores to achieve further improvements. In contrast, larger models exhibit more significant gains in dense reasoning benchmarks: the integration of latent visual memory overcomes bottlenecks in visual reasoning by providing fine-grained visual evidence and semantic knowledge. Notably, this model-agnostic approach, independent of specific model architectures or structures, bolsters the prospects for broad practical application.

\begin{table*}[t]
\centering
\caption{Ablations of latent vision memory invocation and dual vision memory formation. Following~\cite{zhang2025memgen}, ``Random Invocation" denotes that the latent memory is inserted into the output sequence with a certain probability when outputting delimiter symbol tokens, and short or long latent memory is inserted with equal probability. When only utilizing short or long latent memory, we directly skip the formation of the specific memory if  invocation tokens are predicted and continue the process of decoding.}
\setlength{\tabcolsep}{0.9mm}
\resizebox{0.9\textwidth}{!}{
\begin{tabular}{l|cccccccccccc}
\toprule   
\multirow{2}{*}{\textbf{Ablation}} & \multicolumn{3}{c}{\textbf{MMVet}} & \multicolumn{3}{c}{\textbf{MuirBench}} & \multicolumn{3}{c}{\textbf{MV-Math}} & \multicolumn{3}{c}{\textbf{MultiTrust}}  \\ \cmidrule(lr){2-4} \cmidrule(lr){5-7} \cmidrule(lr){8-10} \cmidrule(lr){11-13}
 & Time & Speed & Perf. & Time & Speed & Perf. & Time & Speed & Perf. & Time & Speed & Perf.  \\ \midrule
Vanilla & \textbf{0.76} & \textbf{1.32} & 66.0 & \textbf{3.79} & \textbf{0.26} & 57.4 & \textbf{5.47} & \textbf{0.18} & 18.9 & \textbf{3.62} & \textbf{0.28} & 64.8 \\ \midrule
Random Invocation (25\%) & 0.80 & 1.25 & 69.2 & \underline{3.94} & \underline{0.25} & 59.4 & 8.79 & 0.11 & 29.8 & 6.14 & 0.16 & 69.4 \\
Random Invocation (50\%) & 0.83 & 1.20 & 71.9 & 4.12 & 0.24 & 63.2 & 11.68 & 0.09 & 26.1 & 8.62 & 0.12 & 68.5 \\
Random Invocation (75\%) & 0.86 & 1.16 & 73.6 & 4.27 & 0.23 & 62.7 & 14.78 & 0.07 & 21.9 & 10.11 & 0.10 & 63.7 \\
Full Invocation (100\%) & 0.88 & 1.14 & 73.4 & 4.43 & 0.23 & 56.0 & 17.87 & 0.06 & 17.5 & 13.43 & 0.07 & 62.6 \\ \midrule
Short-term Memory & \underline{0.79} & \underline{1.27} & 71.5 & 4.00 & 0.25 & \underline{65.6} & 7.64 & 0.12 & 29.6 & 4.96 & 0.20 & \underline{73.6} \\
Long-term Memory & 0.81 & 1.23 & \underline{69.4} & 3.95 & 0.25 & 60.2 & \underline{7.61} & \underline{0.12} & \underline{36.1} & \underline{4.80} & \underline{0.21} & 69.8 \\ \midrule
\rowcolor{gray!20} \textbf{Complete VisMem (Ours)} & 0.84 & 1.19 & \textbf{75.1} & 4.10 & 0.24 & \textbf{69.8} & 7.87 & 0.13  & \textbf{41.4} & 5.85 & 0.17 & \textbf{77.0} \\ \bottomrule
\end{tabular}}
\label{tab:ablation_appendix}
\end{table*}

\subsection{Ablation Study}
The vanilla model establishes a baseline characterized by the shortest inference time and highest speed across all benchmarks, yet exhibits the lowest performance. This confirms that latent vision memory is indispensable for enhancing task performance. For the random memory invocation variants, increasing the invocation probability (25\%–100\%) results in longer inference time and reduced speed. Performance peaks at a 75\% probability before declining, indicating that excessive memory invocation impairs efficiency without yielding additional performance benefits. Ablation studies of the short-term and long-term memory components reveal task-specific advantages: the short-term memory component outperforms on MuirBench~\cite{wang2025muirbench} and MultiTrust~\cite{zhang2024multitrust}, while the long-term component demonstrates superior performance on MV-Math~\cite{wang2025mv}. Notably, the complete VisMem framework achieves the highest performance across all benchmarks, validating the value of integrating dual-component vision memory for balanced and robust visual capacities.

 \begin{figure*}[t]
    \centering 
    \begin{subfigure}[b]{0.48\textwidth} 
        \centering
        \includegraphics[width=\textwidth]{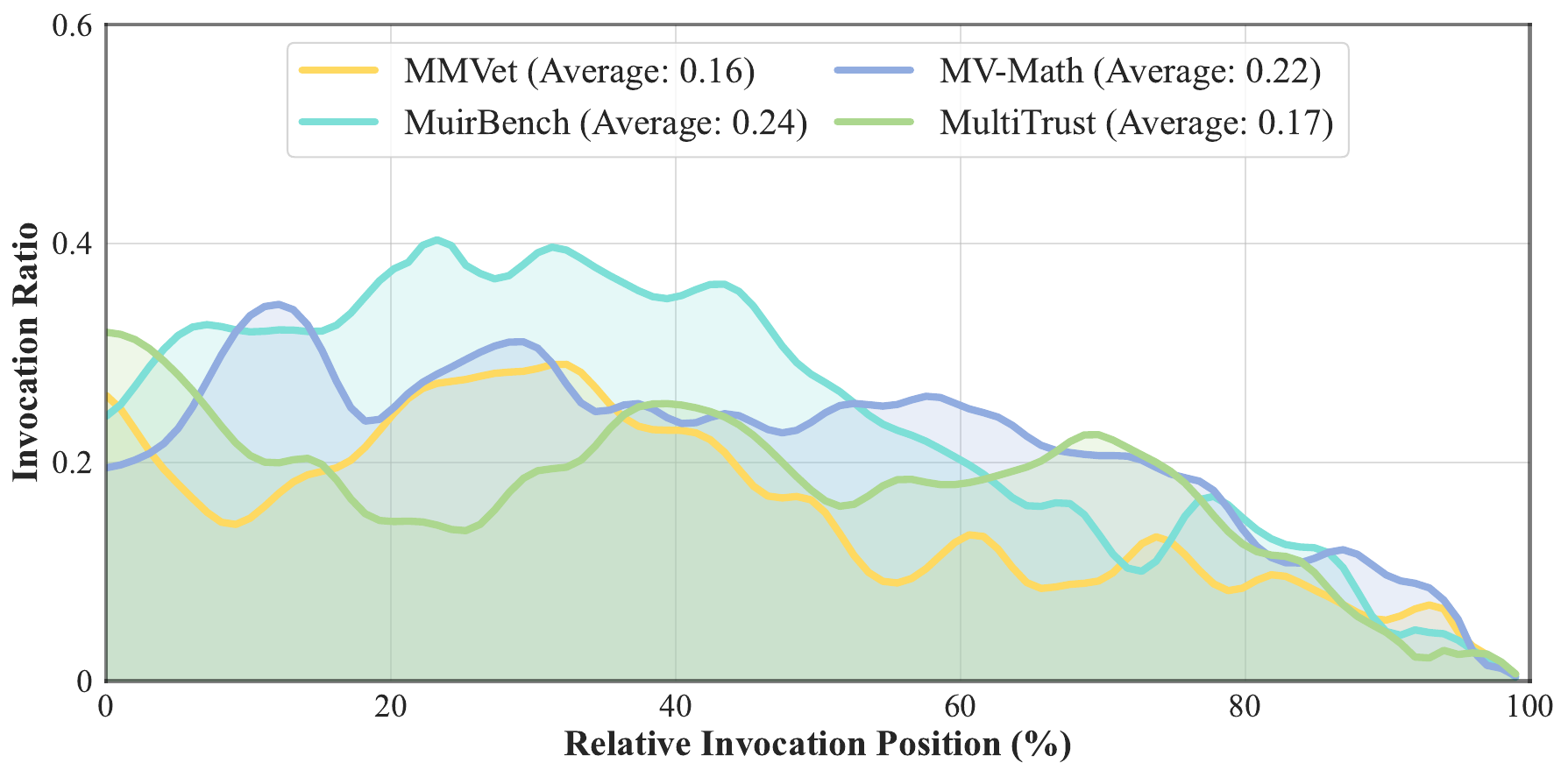} 
        \caption{Short Memory Invocation} 
    \end{subfigure}
    \hfill  
    \begin{subfigure}[b]{0.48\textwidth}  
        \centering
        \includegraphics[width=\textwidth]{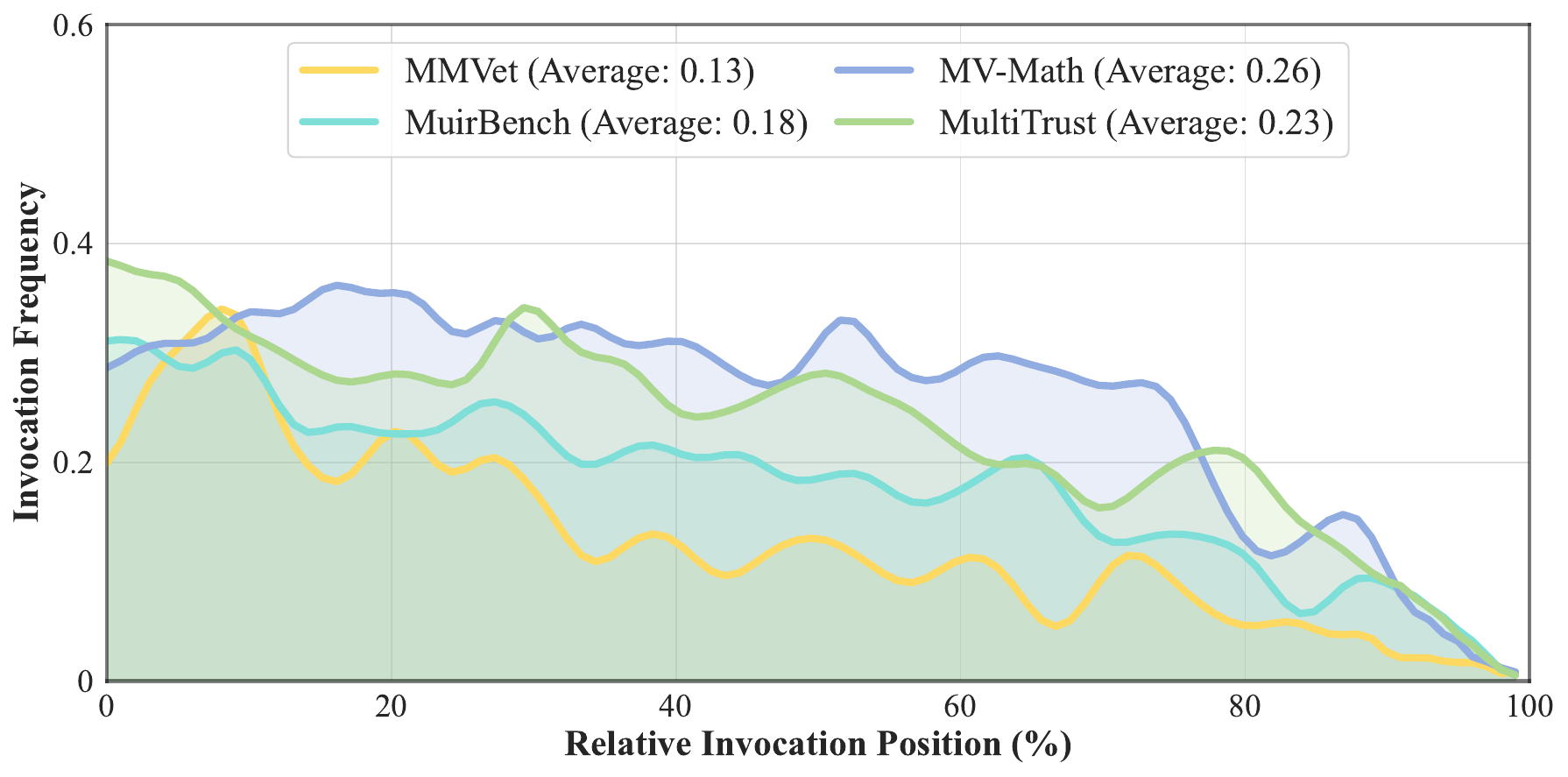}
        \caption{Long Memory Invocation}  
    \end{subfigure}
    \caption{Results of memory invocation ratio and relative position across four benchmarks. The former denotes the proportion of invoked samples to all samples, while the relative position denotes the position in the whole output sequence when the invocation occurred. We apply gaussian smoothing to the curves to highlight their main trends.}
    \label{fig:memory_invocation_appendix} 
\end{figure*}

\begin{figure}[t]
    \centering
    \includegraphics[width=0.8\linewidth]{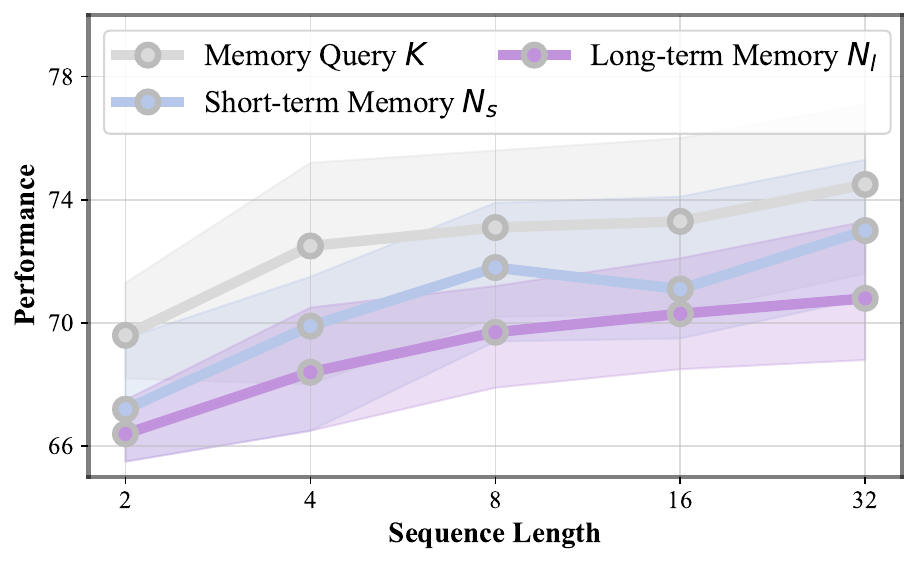}
    \caption{Results of sensitivity analysis on the sequence length of memory query $K$, short- and long-term memory $N_s$ and $N_l$.}
    \label{fig:sequence_length}
\end{figure}

\begin{table}[t]
\centering
\caption{Results of different length of memory query $K$.}
\setlength{\tabcolsep}{0.9mm}
\resizebox{0.46\textwidth}{!}{
\begin{tabular}{c|cccc}
\toprule   
$K$ & \textbf{MMVet} & \textbf{MuirBench}  & \textbf{MV-Math} & \textbf{MultiTrust} \\\midrule
\multicolumn{1}{c}{Vanilla} &  66.0 & 57.4 & 18.9 & 64.8 \\ \midrule
2 & 69.6 & 66.0 & 34.7 & 71.9 \\
4 & 72.5 & 68.9 & 40.6 &  74.8 \\
\rowcolor{gray!20} 8 & 73.1 & 69.8 & \underline{41.1} & 77.0 \\
16 & \underline{73.3} & \underline{70.0} & \textbf{41.4} & \underline{77.7} \\
32 & \textbf{74.5} & \textbf{70.3} & 40.9 & \textbf{78.2} \\ \bottomrule 
\end{tabular}}
\label{tab:query_length}
\end{table}

\begin{table}[t]
\centering
\caption{Results of different length of short latent vision memory $N_s$ and the length of long latent vision memory $N_l$ across four benchmarks.}
\setlength{\tabcolsep}{0.9mm}
\resizebox{0.46\textwidth}{!}{
\begin{tabular}{cc|cccc}
\toprule   
$N_s$ & $N_l$ & \textbf{MMVet} & \textbf{MuirBench}  & \textbf{MV-Math} & \textbf{MultiTrust} \\\midrule
\multicolumn{2}{c}{Vanilla} & 66.0 & 57.4 & 18.9 & 64.8 \\ \midrule
2 & - & 67.2 & 63.7 & 28.2 & 69.3 \\
4 & - & 69.9 & 64.6 & 31.5 & 71.4 \\
8 & - & 71.8 & 65.2 & 33.8 & 73.4 \\
16 & - & 71.1 & 67.8 & 34.0 & 73.3 \\
32 & - & \underline{73.0} & \underline{69.1} & 34.4 & 72.7 \\ \midrule
- & 2 & 66.4 & 60.3 & 29.3 & 71.0 \\
- & 4 & 68.4 & 61.8 & 32.4 & 72.8 \\
- & 8 & 69.7 & 63.0 & 33.5 & 74.2 \\
- & 16 & 70.3 & 63.4 & 34.8 & 74.9 \\
- & 32 & 70.8 & 63.1 & \underline{35.5} & \underline{75.3} \\ \midrule
\rowcolor{gray!20} 8 & 16 & \textbf{75.1} & \textbf{69.8} & \textbf{41.1} & \textbf{77.0} \\
 \bottomrule 
\end{tabular}}
\label{tab:memory_length}
\end{table}

\subsection{Analysis of Latent Vision Memory}
We visualize the invocation ratio and relative invocation position, as presented in \cref{fig:memory_invocation} and \ref{fig:memory_invocation_appendix}: the former illustrates benchmark-specific differences between the two memory components, while the latter depicts type-specific variations across the four benchmarks. In addition, as reported in \cref{tab:muirbench} and \ref{tab:logic_vista}, the short- and long-term latent visual memory components exhibit task-specific advantages for different visual sub-tasks. For instance, the short-term memory provides supplementary visual information to support enhanced visual understanding, such as counting, grounding, and visual retrieval. By contrast, the long-term memory encodes contextualized semantic knowledge, which strengthens complex visual reasoning. These results reveal that our proposed VisMem dynamically adjusts invocation position and frequency according to task characteristics, thereby balancing efficiency and performance.

\subsection{Sensitive Analysis of Sequence Lengths}
We conduct an analysis on MMVet~\cite{yu2024mmvet} focused on the lengths of three key sequences: the memory query $K$, the short-term latent visual memory $N_s$, and the long-term latent visual memory $N_l$. It is observed that as the lengths of these three sequences increase from 2 to 32, model performance improves accordingly, but this is accompanied by increased computational costs.

\subsection{Inference Efficiency}
As presented in \cref{tab:efficiency} and the bubble plots in \cref{fig:efficiency}, we compare the average inference time, average inference speed, and task performance across the four benchmarks. Our approach achieves an optimal performance-efficiency balance, with minimal additional time overhead. For instance, image-level paradigms exhibit nearly twice the inference time of the vanilla model, resulting in significant latency and substantial inference overhead. In contrast, our VisMem introduces only controllable computational latency increments, ranging from 8.2\% to 43.8\% relative to the vanilla model, which are on par with those of other direct training and token-level paradigms.

\begin{table*}[t]
\centering
\caption{Average inference time per sample (seconds), average inference speed (samples / seconds), and task performances across four benchmarks on various methods. Perf. indicates Performance.}
\setlength{\tabcolsep}{0.9mm}
\resizebox{0.75\textwidth}{!}{
\begin{tabular}{l|cccccccccccc}
\toprule   
\multirow{2}{*}{\textbf{Method}} & \multicolumn{3}{c}{\textbf{MMVet}} & \multicolumn{3}{c}{\textbf{MuirBench}} & \multicolumn{3}{c}{\textbf{MV-Math}} & \multicolumn{3}{c}{\textbf{MultiTrust}}  \\ \cmidrule(lr){2-4} \cmidrule(lr){5-7} \cmidrule(lr){8-10} \cmidrule(lr){11-13}
 & Time & Speed & Perf. & Time & Speed & Perf. & Time & Speed & Perf. & Time & Speed & Perf.  \\ \midrule
Vanilla~\cite{bai2025qwen2} & \underline{0.76} & \underline{1.32} & 66.0 & \textbf{3.79} & \textbf{0.26} & 57.4 & \textbf{5.47} & \textbf{0.18} & 18.9 & \textbf{3.62} & \textbf{0.28} & 64.8 \\ \midrule
SFT & \textbf{0.75} & \textbf{1.33} & 67.5 & 3.82 & 0.26 & 58.7 & 6.35 & 0.16 & 22.8 & 3.68 & 0.27 & 67.0 \\
Visual-RFT~\cite{liu2025visual} & 0.76 & 1.32 & 70.5 & \underline{3.81} & \underline{0.26} & 62.9 & \underline{5.66} & \underline{0.17} & 26.5 & \underline{3.65} & \underline{0.27} & 70.7 \\
VLM-R1~\cite{shen2025vlm} & 0.77 & 1.30 & \underline{73.0} & 3.83 & 0.26 & 63.8 & 7.88 & 0.13 & 34.6 & 3.69 & 0.27 & 69.9 \\
Vision-R1~\cite{huang2025vision} & 0.77 & 1.30 & 71.7 & 3.83 & 0.26 & \underline{64.0} & 8.42 & 0.12 & 38.7 & 3.71 & 0.27 & 72.6 \\
PAPO~\cite{wang2025perception} & 0.76 & 1.32 & 69.8 & 3.81 & 0.26 & 56.7 & 6.74 & 0.15 & 34.8 & 3.68 & 0.27 & 67.7 \\ \midrule
Sketchpad~\cite{hu2024visual} & 2.39 & 0.42 & 64.5 & 8.90 & 0.11 & 52.8 & 9.10 & 0.11 & 24.6 & 5.47 & 0.18 & 66.2 \\
GRIT~\cite{fan2025grit} & 0.80 & 1.25 & 67.8 & 4.07 & 0.25 & 51.0 & 8.45 & 0.12 & 22.4 & 4.06 & 0.25 & 67.3 \\
PixelReasoner~\cite{su2025pixel} & 1.45 & 0.69 & 67.1 & 7.34 & 0.14 & 60.5 & 9.96 & 0.10 & 25.9 & 5.60 & 0.18 & 69.9 \\
DeepEyes~\cite{zheng2025deepeyes} & 3.21 & 0.31 & 70.5 & 8.46 & 0.12 & 63.0 & 11.72 & 0.09 & 31.5 & 6.14 & 0.16 & 72.6 \\
OpenThinkImg~\cite{su2025openthinkimg} & 3.68 & 0.27 & 71.6 & 8.69 & 0.12 & 61.7 & 10.38 & 0.10 & 28.0 & 6.43 & 0.16 & \underline{74.0} \\ \midrule
Scaffold~\cite{lei2024scaffolding} & 0.83 & 1.20 & 67.0 & 4.35 & 0.23 & 52.9 & 7.01 & 0.14 & 21.0 & 3.88 & 0.26 & 68.5 \\
ICoT~\cite{gao2025interleaved} & 0.97 & 1.15 & 67.9 & 4.57 & 0.22 & 57.0 & 8.94 & 0.11 & 30.8 & 4.20 & 0.24 & 69.1 \\
MINT-CoT~\cite{chen2025mint} & 0.81 & 1.23 & 69.5 & 4.18 & 0.24 & 58.9 & 7.89 & 0.13 & \underline{39.2} & 4.03 & 0.25 & 71.4 \\
VPT~\cite{yu2025introducing} & 2.98 & 0.34 & 70.8 & 9.63 & 0.10 & 63.5 & 9.59 & 0.10 & 34.7 & 5.79 & 0.17 & 64.7 \\ \midrule
Mirage~\cite{yang2025machine} & 0.86 & 1.16 & 71.8 & 4.02 & 0.25 & 59.0 & 7.71 & 0.13 & 35.4 & 3.82 & 0.26 & 66.1 \\
\rowcolor{gray!20} \textbf{VisMem (Ours)} & 0.84 & 1.19 & \textbf{75.1} & 4.10 & 0.24 & \textbf{69.8} & 7.87 & 0.13 & \textbf{41.4} & 3.85 & 0.26 & \textbf{77.0} \\  \bottomrule
\end{tabular}}
\label{tab:efficiency}
\end{table*}

